\documentclass[article,letterpaper, 10 pt, conference]{ieeeconf}  

\IEEEoverridecommandlockouts                              

\usepackage{graphicx,subfig}

\usepackage{hyperref}
\usepackage[sorting=none, style=ieee]{biblatex}
\usepackage{booktabs}  
\usepackage{threeparttable} 
\usepackage{makecell}
\usepackage{graphicx}
\usepackage{multicol}
\usepackage{multirow}
\usepackage{feyn}
\usepackage{hyperref}
\usepackage{amssymb}
\usepackage{multirow}
\usepackage{tabularx}
\usepackage{amsmath}
\usepackage{longtable}
\usepackage{pifont}

\usepackage{array}
\usepackage{colortbl}

\usepackage{url}

\definecolor{url_color}{RGB}{124,214,207}  

\hypersetup{
        colorlinks=true,
	colorlinks=true,
	linkcolor=cyan,
	filecolor=blue,      
	urlcolor= black,
	citecolor=green,
}

\title{\LARGE \bf
Monocular 3D \textcolor{black}{Lane Detection} for Autonomous Driving: Recent Achievements, Challenges, and Outlooks
}

\author{Fulong Ma, Weiqing Qi, Guoyang Zhao, Linwei Zheng, Sheng Wang, Yuxuan Liu, Ming Liu, and Jun Ma
\thanks{Fulong Ma, Weiqing Qi, Guoyang Zhao, Linwei Zheng, Sheng Wang, and Ming Liu are with The Hong Kong University of Science and Technology (Guangzhou), Guangzhou, China (email:{\{fmaaf, wqiad, gzhao492, lzhengad, swangei, eelium\}@connect.hkust-gz.edu.cn)}.}
\thanks{Yuxuan Liu and Jun Ma are with The Hong Kong University of Science and Technology, Hong Kong SAR, China (email: yliuhb@connect.ust.hk; jun.ma@ust.hk).}
}

\makeatother

\begin{document}
\maketitle
\thispagestyle{empty}
\pagestyle{empty}

\begin{abstract}
3D lane detection is essential in AD as it extracts structural and traffic information from the road in 3D space, aiding autonomous vehicles in logical, safe, and comfortable path planning and motion control. Given the cost of sensors and the advantages of visual data in color information, 3D lane detection based on monocular vision is an important research direction in the realm of AD that increasingly gains attention in both industry and academia. Nevertheless, recent advancements in visual perception seem inadequate for the development of fully reliable 3D lane detection algorithms, which also hampers the progress of vision-based fully autonomous vehicles. 
We believe that it still leaves an open and interesting problem for improvement in
3D lane detection algorithms for autonomous vehicles using visual sensors, and significant enhancements are essentially required. 
This review summarizes and analyzes the current state of achievements in the field of 3D lane detection research. It covers all current monocular-based 3D lane detection processes, discusses the performance of these cutting-edge algorithms, analyzes the time complexity of various algorithms, and highlights the main achievements and limitations of ongoing research efforts. The survey also includes a comprehensive discussion of available 3D lane detection datasets and the challenges that researchers encounter but have not yet resolved. Finally, our work outlines future research directions and invites researchers and practitioners to join this exciting field.


\end{abstract}

\begin{figure*}[h]
    \setlength{\abovecaptionskip}{0pt}
    \setlength{\belowcaptionskip}{0pt}
    \centering
    \includegraphics[width=1.0\linewidth]{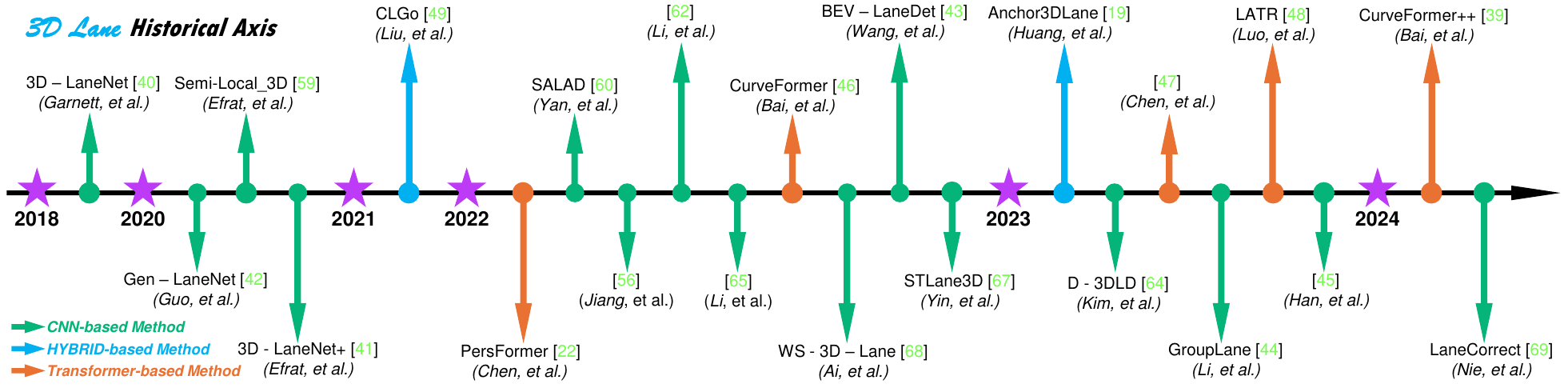}
    \captionsetup{font={footnotesize}}
    \caption{
    Chronological overview of 3D lane detection based on monocular images. Green arrows represent methods based on CNNs, orange arrows represent methods based on Transformers, and blue arrows represent hybrid architecture methods.
}
    \label{timeline}
\end{figure*}
\section{INTRODUCTION}

In recent advancements fueled by artificial intelligence, AD technology has seen rapid evolution, progressively reshaping the paradigms of human transportation. Equipped with an array of sensors, autonomous vehicles emulate human sensory capabilities such as vision and hearing to perceive their surroundings and interpret traffic scenarios for safe navigation \cite{wei2021multi}. Key among these sensors are LiDAR, high-resolution cameras, millimeter-wave radars, and ultrasonic radars \cite{zou2021comparative,ma2022automatic}, which facilitate feature extraction and object classification combined with high-precision mapping to discern obstacles and the vehicular traffic landscape \cite{zhao2024curbnet}. Visual sensors, which are most extensively utilized in autonomous vehicles, serve as the primary means for environmental perception, encompassing lane detection, traffic light analysis, road sign detection and recognition, vehicle tracking, pedestrian detection, and short-term traffic prediction \cite{chen2021data}. The processing and understanding of visual scenes in AD, which includes analysis of traffic lights, identification of traffic signs, lane detection, and the detection of nearby pedestrians and vehicles, underpin more robust and safer commands for maneuvers such as steering, overtaking, lane changes, or braking \cite{usman2020speed,ma2024dataset}. This integration of sensory data and environmental understanding seamlessly transitions into the domain of scene understanding in AD, which is pivotal for advancing vehicular autonomy and ensuring road safety.

Scene understanding represents one of the most challenging aspects within the domain of AD \cite{ma2023self}. Without the capability for comprehensive scene understanding, navigating an autonomous vehicle safely through traffic lanes can be as daunting as navigating the world blindfolded for humans. Lane detection, in particular, stands out as a pivotal and challenging task within the realm of scene comprehension \cite{xu2023centerlinedet}. Lanes serve as the most common traffic elements on roads, acting as crucial markers that segment the roadways to ensure the safe and efficient passage of vehicles \cite{tang2021review}. Lane detection technology, which automatically identifies road markings, is indispensable; autonomous vehicles lacking this capability could lead to traffic congestion and even severe collisions, thereby compromising passenger safety. Thus, lane detection plays a critical role in the ecosystem of AD. Unlike typical objects, lane markings occupy only a slender portion of the road scene and are widely distributed, making them uniquely challenging to detect \cite{sun2019accurate}. The task is further complicated by a variety of lane markings, insufficient lighting, obstructions, and interference from similar textures, which are common in many driving scenarios, thus exacerbating the intrinsic challenges associated with lane detection.

Lane detection methods based on monocular vision can be primarily categorized into traditional manual feature approaches and deep learning-based methods \cite{ma2018multiple}. Early endeavors primarily focused on extracting low-level manual features \cite{he2004color,kim2008robust}, such as edges and color information. However, these approaches often involve complex feature extraction and post-processing designs, and exhibit limited robustness in dynamically changing scenes. Traditional lane detection algorithms based on manual feature extraction commence by identifying characteristics such as color, texture, edges, orientation, and shape of the lane lines, followed by constructing detection models that approximate the lane markings as straight or higher-order curved lines \cite{zhou2010novel}. Nevertheless, due to the lack of distinguishing features and poor adaptability to dynamic environments, traditional methods based on manual features are typically less reliable and more computationally expensive \cite{wang2004lane}. With the rapid evolution of deep learning, significant strides have been made in the fields of image classification, object detection, and semantic segmentation within computer vision, bringing innovative perspectives to the research of lane detection \cite{wei2022row}. Deep neural networks (DNNs), rooted in deep learning, demonstrate profound capability in feature extraction from image data, with Convolutional Neural Networks (CNNs) being the most extensively applied. CNNs represent a specialized category of DNNs, characterized by multiple convolutional layers and bases, making them particularly suitable for processing structured data like visual images, and providing efficient feature extraction for various subsequent tasks \cite{xu2022rngdet}. In the context of lane detection, this translates to utilizing deep CNNs to extract high-level features in real-time frames, which are then processed by the model to accurately determine the positions of lane lines \cite{huang2023anchor3dlane}.

\subsection{Background and Related Works}

Due to advancements in deep learning technologies, researchers have developed numerous strategies that have significantly simplified, expedited, and enhanced the task of lane detection \cite{tabelini2021keep}. Concurrently, as deep learning becomes increasingly widespread and as new concepts continuously emerge \cite{ma20233d}, the domain of lane detection has seen a further specialization and refinement in its approaches. Reflecting on the mainstream research trajectories within this domain, camera-based lane detection methods can be principally segmented into 2D and 3D lane detection paradigms \cite{chen2022persformer}.

\textbf{2D lane detection.} 
In \cite{jin2022eigenlanes,liu2021end}, these works aim to accurately delineate the shape and position of lanes within images, primarily employing four distinct approaches: segmentation-based, anchor-based, keypoint-based, and curve-based strategies.
(1) Segmentation-based approaches frame 2D lane detection as a pixel-level classification challenge, generating lane masks \cite{zheng2021resa,neven2018towards,pan2018spatial,zheng2021resa}. These methods cultivate lane groupings through the exploration of effective semantic features and subsequent post-processing, although they are marked by their high computational costs.
(2) Anchor-based approaches are lauded in 2D lane detection for their simplicity and efficiency, typically utilizing linear anchor points to regress positional offsets relative to targets \cite{li2019line,tabelini2021keep,zheng2022clrnet}. To circumvent the constraints of linear anchors, 
various lane candidates are generated by an intrinsic lane space in \cite{jin2022eigenlanes}.
Heuristically designed row anchors classify row pixels as lanes \cite{yoo2020end,liu2021condlanenet}, further evolved into mixed (row and column) anchors in \cite{qin2022ultra} to alleviate positioning errors for side lanes. Moreover, this method significantly enhances inference speed.
(3) Keypoint-based approaches offer a more flexible and sparse modeling of lane positions \cite{ko2021key,qu2021focus,wang2022keypoint}, initiating with the estimation of point locations, followed by associating keypoints belonging to the same lane using varying schemes. The primary strategy in \cite{su2021structure,tabelini2021keep} is to predict 2D lanes by representing lane structures with predefined keypoints and regressing offsets between sampled and predefined points. Despite promising results, these methods lack the flexibility to adapt to complex lane configurations due to their fixed point design.
(4) Curve-based approaches \cite{tabelini2021polylanenet,van2019end,liu2021end}focus on fitting lane lines through various curve equations and specific parameters, transforming 2D lane detection into a curve parameter regression challenge by detecting start and end points alongside curve parameters. Although there have been promising advancements in 2D lane detection, a significant gap remains between the 2D outcomes and the real-world application requirements, notably the precise 3D positioning.

\textbf{3D lane detection.} Due to the inherent lack of depth information in 2D lane detection, projecting these detections into 3D space can result in inaccuracies and reduced robustness \cite{bai2024curveformer++}. Consequently, numerous researchers have shifted their focus towards lane detection within the 3D domain. Deep learning-based 3D lane detection methodologies primarily bifurcate into CNN-based methods, Transformer-based methods, \textcolor{black}{and hybrid methods.}

Representative CNN-based methods predominantly include 3D-LaneNet \cite{garnett20193d}, which proposes a dual-path architecture utilizing Inverse Perspective Mapping (IPM) to transpose features and detect lanes through vertical anchor regression. 3D-LaneNet+ \cite{efrat20203d} segments bird’s eye view (BEV) features into non-overlapping units, addressing the limitations of anchor orientation through lateral offsets, angles, and elevation shifts relative to the unit's center. GenLaneNet \cite{guo2020gen} pioneers the use of a fictitious top-view coordinate system for better feature alignment and introduces a two-stage framework to decouple lane segmentation and geometric encoding. BEVLaneDet \textcolor{black}{\cite{wang2023bev} employs} a virtual camera to ensure spatial consistency and adapts to more complex scenarios through a keypoint-based representation of 3D lanes. 
\textcolor{black}{Anchor3DLane \cite{huang2023anchor3dlane} regresses 3D lanes directly from image features based on 3D anchors, significantly reducing computational overhead.} 
GroupLane \cite{li2023grouplane} innovates within the BEV by introducing a row-based classification method, which accommodates lanes in any direction and interacts with feature information within instance groups.
\textcolor{black}{DecoupleLane \cite{han2023decoupling} divides 3D lane detection into two parts: curve modeling and ground height regression. By representing lanes in the BEV space and modeling ground height separately, this method effectively addresses the fluctuations caused by uneven road surfaces.}

\textcolor{black}{Transformer-based methods include PersFormer} \cite{chen2022persformer}, which constructs dense BEV queries using offline camera poses, unifying 2D and 3D lane detection under a Transformer-based framework. 
CurveFormer \cite{bai2022curveformer} leverages sparse query representations and cross-attention mechanisms within Transformers to effectively regress polynomial coefficients of 3D lanes. 
\textcolor{black}{In \cite{chen2023efficient}, Chen \textit{et al.} introduce an efficient transformer for 3D lane detection that jointly learns BEV and lane representations. Unlike methods using IPM for BEV transformation before lane detection, the model uses a decomposed cross-attention mechanism to learn both representations with supervision. }
LATR \cite{luo2023latr}, building on CurveFormer’s query anchor modeling, constructs lane-aware query generators and dynamic 3D ground position embeddings. CurveFormer++\cite{bai2024curveformer++} proposes a single-stage Transformer detection method that does not require image feature view transformation and directly infers 3D lane detection results from perspective image features.

\textcolor{black}{A series of methods also utilize hybrid network architectures, combining CNN and Transformer for 3D lane detection.
CLGo \cite{liu2022learning} proposes a two-stage framework. The first stage uses a CNN backbone for feature extraction and a transformer-encoder that aggregates nonlocal relations among spatial features. Additionally, it estimates the camera pose by introducing an auxiliary 3D lane task and geometric constraints. In the second stage, the estimated camera pose is used to generate a BEV image, which is then used to predict 3D lanes.
Anchor3DLane \cite{huang2023anchor3dlane} uses a CNN backbone and Transformer layer to extract features and directly predict 3D lanes from front-view images, bypassing the need for inverse perspective mapping. It defines 3D lane anchors as rays and projects them onto the feature map to capture necessary structural and contextual information for accurate 3D lane detection.}

\subsection{Challenges and Motivation}
\label{questions}
Accurately estimating the 3D position of lane markings based on monocular vision presents numerous challenges. Firstly, real-world data for 3D lane detection is highly variable due to the diversity of weather \cite{kuang2020real}, road types, markings, and environmental conditions, which complicates the training of models that can generalize well across different scenarios \cite{campbell2010autonomous,zhang2023perception}. Furthermore, processing 3D data for lane detection requires substantial computational resources, which is particularly critical in AD applications where low latency is imperative. Reducing the processing time of algorithms without compromising their performance is also a significant challenge in such real-time applications.

Considering the aforementioned challenges and the importance of \textcolor{black}{vision sensor-based 3D lane detection} in accurate traffic scene understanding and parsing, we have accumulated existing research contributions and outcomes in this survey. 
\textcolor{black}{Furthermore, before beginning this survey, we have posed the following questions:}
1) Do existing datasets possess the potential for 3D lane detection in complex visual scenes? 
2) What are the \textcolor{black}{ inference speeds }of current methods, and can these methods meet the real-time requirements of autonomous vehicles? 
3) Can current methods effectively perform 3D lane detection in complex visual scenes containing uncertainties such as fog and rain?

\subsection{Contributions}
This survey takes a step ahead by critically examining the recent state-of-the-art in 3D lane detection techniques, and makes the following main contributions to the community:
\begin{enumerate}[]
\item A discussion and critical analysis of the most relevant papers and datasets that have received significant attention in the field of 3D lane detection in recent years.
\item A comprehensive introduction to 3D lane detection techniques, defining the generic pipeline and explaining each step individually. This encourages newcomers in the field to quickly grasp prior knowledge and contributions from previous research, especially in the context of AD.
To the best of our knowledge, this is the first survey on camera-based 3D lane detection.
\item A performance study of current state-of-the-art methods, considering their computational resource requirements and the platforms for which these methods are developed. 

\item A reasoned derivation of future research guidelines based on the analyzed literature, identifying open problems and challenges in this domain, as well as research opportunities that can be explored effectively to address them.
\end{enumerate}

\subsection{Review Methodology}
The research works discussed in this survey were retrieved using different keywords such as 3D lane detection in autonomous vehicles, vision-based 3D lane detection, and learning-based 3D lane detection. Most of the retrieved papers are directly relevant to the research topic, but there are also some exceptions, such as multi-modal methods \cite{luo2022m,bai2018deep}, and point cloud-based approaches \cite{jung2018real}, which are less related to the theme of this survey. Additionally, the aforementioned keywords were searched in multiple repositories, including Web of Science and Google Scholar, to ensure the retrieval of relevant content. Inclusion criteria ensured that a paper was recognized by experts in AD, based on factors such as citation count or the impact of prior work.
It is worth mentioning that in the existing literature, no work on monocular 3D lane detection based on traditional methods was found. This may be because, unlike single-camera 2D lane detection, which only requires identifying pixels belonging to the lane in a 2D image, monocular 3D lane detection requires determining the 3D position information of the lane in the 3D space using the 2D image. Without the assistance of distance measuring sensors such as LiDAR or prediction through deep learning, it is difficult to achieve.






The remainder of this paper is structured as follows: Section \ref{lane_detection_in_ad} provides an
overview of related works for 3D lane detection in AD. 
Section \ref{performance_evaluation} introduces performance evaluation of 3D lane detection. In section \ref{datasetes}, commonly used datasets are introduced.
Next, in section \ref{challenges_and_directions}, we introduce the current challenges and prospects for future 3D lane detection.
Finally, Section \ref{Conclusion} concludes the paper.


\begin{figure}[t]
    \setlength{\abovecaptionskip}{0pt}
    \setlength{\belowcaptionskip}{0pt}
    \centering
    \includegraphics[width=1.0\linewidth]{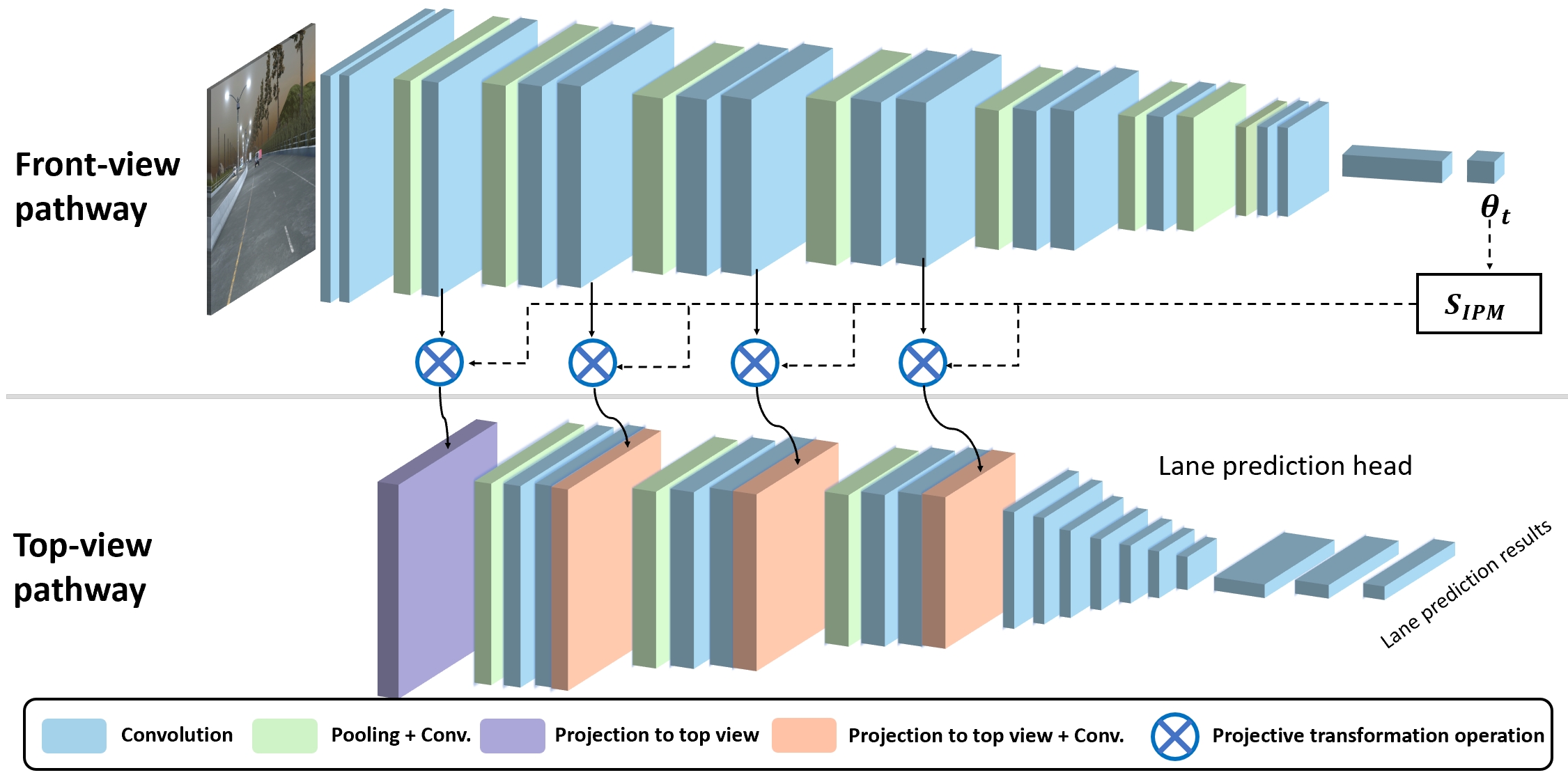}
    \captionsetup{font={footnotesize}}
    \caption{
    An overview of the \textcolor{black}{3D-LaneNet \cite{garnett20193d}} architecture. Information is processed in two parallel streams or pathways: the image-view pathway and the top-view pathway. This is called the dual-pathway backbone.}
    \label{3d_lanenet}
\end{figure}

\begin{figure}[h]
    \setlength{\abovecaptionskip}{0pt}
    \setlength{\belowcaptionskip}{0pt}
    \centering
    \includegraphics[width=1.0\linewidth]{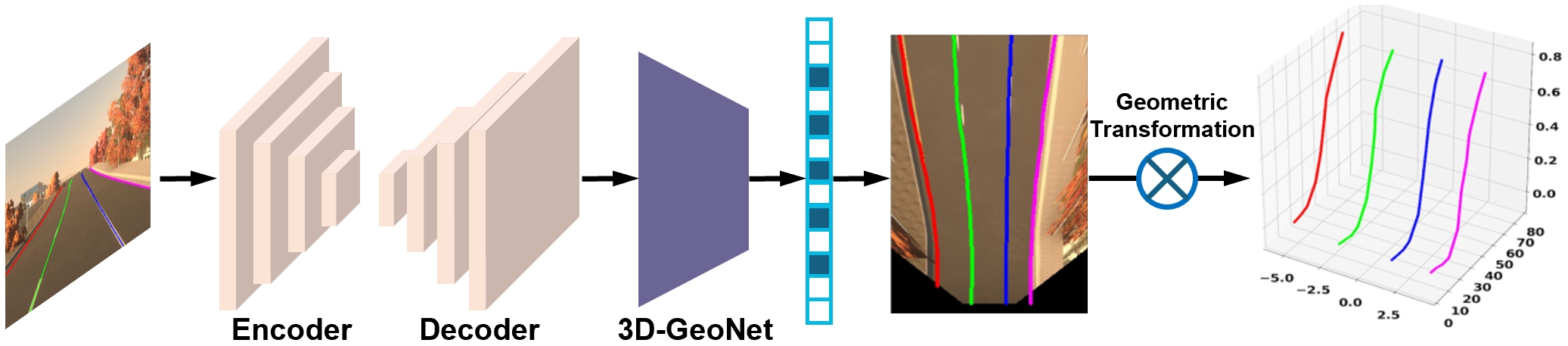}
    \captionsetup{font={footnotesize}}
    \caption{
    The procedures of \textcolor{black}{Gen-LaneNet \cite{guo2020gen}} involve encoding an input image into deep features using a segmentation backbone, which are then decoded into a lane segmentation map. The 3D-GeoNet then focuses on geometry encoding and predicts intermediate 3D lane points represented in top-view 2D coordinates and real heights. Finally, a geometric transformation converts the network output into real-world 3D lane points.}
    \label{gen_lanenet}
\end{figure}

\section{\textcolor{black}{MONOCULAR} 3D LANE DETECTION IN AD}
\label{lane_detection_in_ad}

With the rapid development of AD and deep learning, learning-based monocular lane detection has received increasing attention from both the industry and academia. 
In the field of monocular lane detection, early work mainly focused on 2D lane detection. 
As AD technology matures and the demand for cost reduction increases, higher requirements are placed on lane detection, aiming to predict the 3D information of lane lines from a single image.Therefore, starting from 2018, there have been successive works on monocular 3D lane detection. As shown in Fig. \ref{timeline}, the figure provides a chronological overview of the monocular 3D lane detection algorithms. It can be seen that with the passage of time, more and more works have emerged, indicating that this field is becoming increasingly popular. In this figure, the green arrows represent CNN-based methods, the orange arrows represent Transformer-based methods, \textcolor{black}{and the blue arrows represent methods with hybrid architecture.}

\begin{figure}[h]
    \setlength{\abovecaptionskip}{0pt}
    \setlength{\belowcaptionskip}{0pt}
    \centering
    \includegraphics[width=1.0\linewidth]{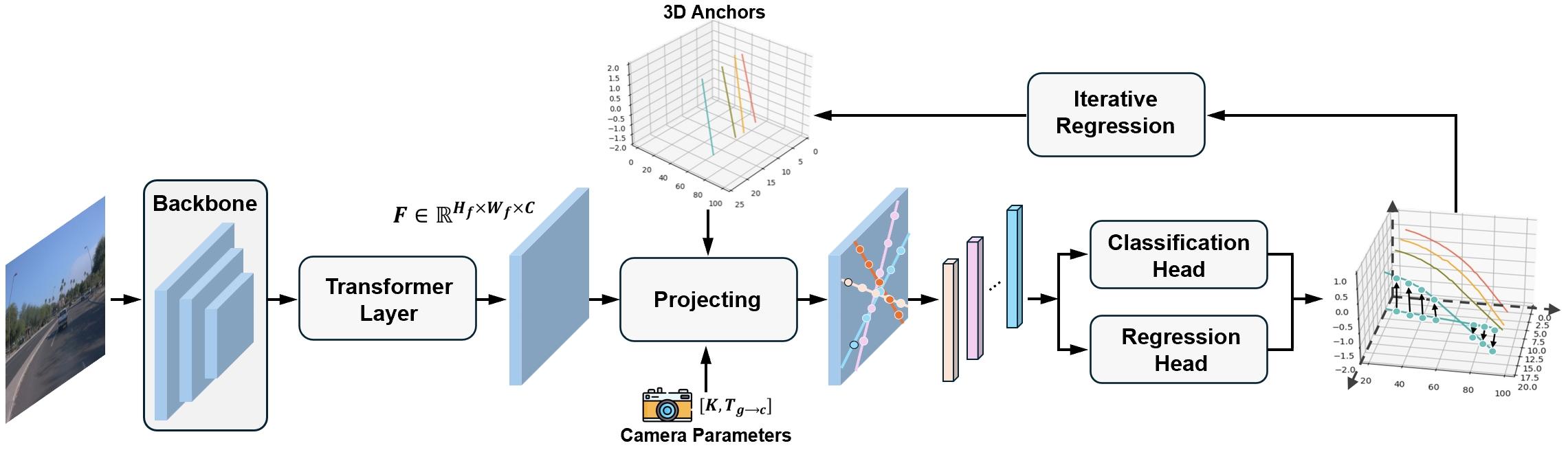}
    \captionsetup{font={footnotesize}}
    \caption{
    The architecture of \textcolor{black}{Anchor3DLane \cite{huang2023anchor3dlane}}. Utilizing a front-view input image, we employ a CNN backbone and a Transformer layer to extract the visual feature $F$ initially. Subsequently, 3D anchors are projected to sample their features from $F$ based on camera parameters. Following this, a classification head and a regression head are utilized to make the final predictions. The lane predictions can also be used as new 3D anchors for iterative regression.}
    \label{anchor3dlane}
\end{figure}

Among these methods, 3D-LaneNet \cite{garnett20193d} is a pioneering work in the field of monocular 3D lane detection.
The overview of 3D-LaneNet is depicted in the Fig. \ref{3d_lanenet}.
3D-LaneNet introduces a network that can directly predict 3D lane information in road scenes from monocular images. This work is the first to solve the task of 3D lane detection using on-board monocular vision sensor. 
3D-LaneNet introduces two novel concepts: intra-network feature map IPM and anchor-based lane representation. The intra-network IPM projection facilitates dual representation information flow in both front view and bird's-eye view. 
\textcolor{black}{The anchor-based lane output representation supports an end-to-end training approach, which has demonstrated a significant impact on subsequent work in this field. Many later studies have adopted the anchor-based lane representation.
Although 3D-LaneNet has shown promising results in 3D lane detection, its limitations lie in using an inappropriate coordinate frame that misaligns lane points with visual features, and in the end-to-end learning framework where geometric encoding is easily affected by changes in image appearance, thus requiring more training data, which limits its practical application.}
\textcolor{black}{Inspired by 3D-LaneNet and its drawbacks, Guo \textit{et al.} propose Gen-LaneNet \cite{guo2020gen}, a generalized and scalable framework for 3D lane detection, as shown in Fig. \ref{gen_lanenet}. Compared to 3D-LaneNet, Gen-LaneNet is still a unified framework that addresses image encoding, spatial transformation of features, and 3D curve extraction within a single network. However, it has major differences in two aspects: a geometric extension to the lane anchor design and a scalable two-stage network that decouples the learning of image encoding and 3D geometry reasoning. Additionally, this paper introduces a highly realistic synthetic image dataset with rich visual variations for developing and evaluating 3D lane detection methods.
Although the new geometry-guided lane anchor representation has better generalization ability to unobserved scenes, it is still limited to long lanes that are roughly parallel to the direction of travel of the ego vehicle.}
\textcolor{black}{Based on Gen-LaneNet, Jiang \textit{et al.} propose Att-Gen-LaneNet \cite{jiang2022robust}, which incorporates two attention mechanisms: the Efficient Channel Attention (ECA) \cite{wang2020eca} attention mechanism and the Convolutional Block Attention Module (CBAM) \cite{woo2018cbam} attention mechanism to enhance algorithm performance, ultimately achieving an improvement in performance compared to Gen-LaneNet.}
\textcolor{black}{Subsequently, Efrat \textit{et al.} propose a tile representation to expand support for lane topologies.
In the two works proposed by Efrat et al. \cite{efrat2020semi,efrat20203d}, a semi-local tile-based representation is employed, which captures local lane structures and road geometrical features, along with a learning-based method that clusters local lane segments into complete 3D lane curves. The difference is that in \cite{efrat2020semi}, uncertainty estimation is introduced into 3D lane line detection for the first time to reflect the detection noise. Such methods still have certain limitations when dealing with complex road curvatures and geometries. Despite using a semi-local tile representation, performance may degrade in handling some extreme cases, such as highly curved roads.}
\textcolor{black}{Instead of utilizing the ground truth camera pose, Liu \textit{et al.} propose CLGO \cite{liu2022learning}, a hybrid architecture method that combines CNN and Transformer, can learn camera pose online. This method follows a two-stage framework. In the first stage, camera pose is estimated by introducing an auxiliary 3D lane task and geometric constraints to improve pose estimation. In the second stage, the estimated camera pose is used to generate top-view images, which are then used for accurate 3D lane prediction.
It is worth mentioning that, unlike anchor-based representations, this method represents lane lines using polynomials.
The advantage of this method is that it can estimate camera parameters in real-time and reduce errors. However, the drawback is that it introduces an additional camera pose estimation step, which increases the system's complexity.
Another hybrid architecture method is the Anchor3DLane \cite{huang2023anchor3dlane} framework, which directly defines anchor points in 3D space and directly regresses 3D lanes from the front view without introducing a top view, as shown in Fig. \ref{anchor3dlane}. The authors also propose a multi-frame extension of Anchor3DLane to exploit well-aligned temporal information and further improve performance. Furthermore, a global optimization method is developed to fine-tune the lane lines by utilizing the lane iso-width property.
}

\begin{figure}[t]
    \setlength{\abovecaptionskip}{0pt}
    \setlength{\belowcaptionskip}{0pt}
    \centering
    \includegraphics[width=1.0\linewidth]{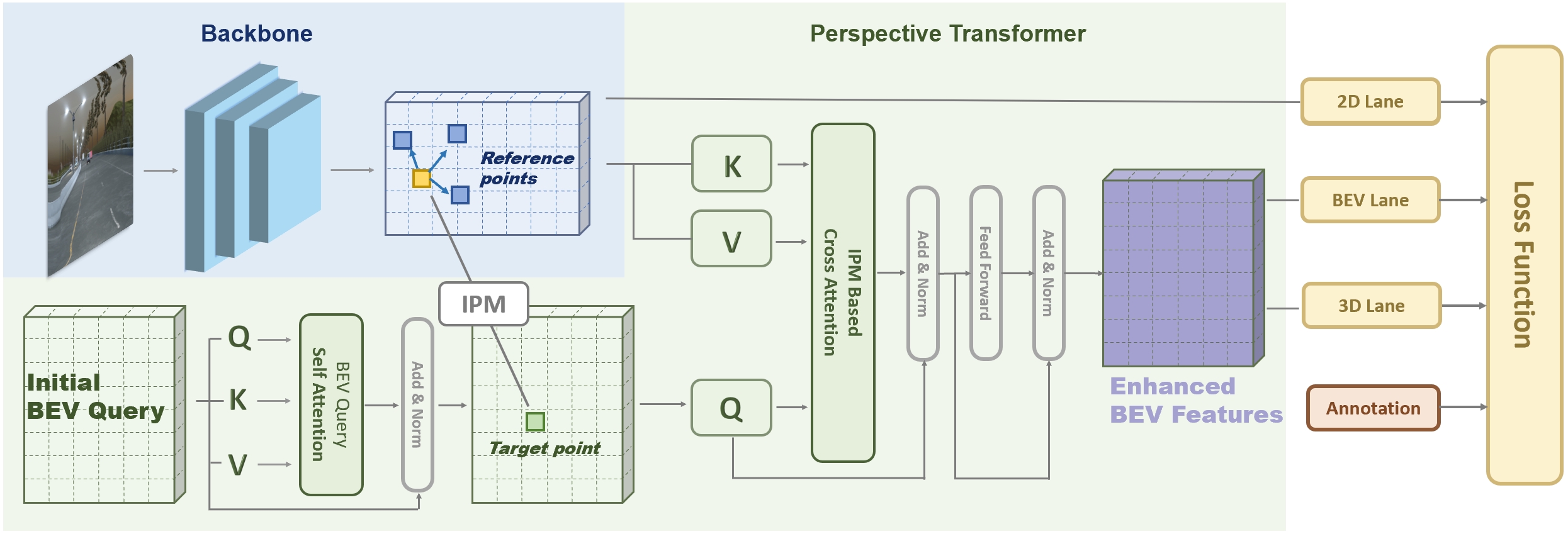}
    \captionsetup{font={footnotesize}}
    \caption{The pipeline of \textcolor{black}{PersFormer \cite{chen2022persformer}}. The key is to understand the spatial feature transformation from the front view to the BEV space, aiming to enhance the representativeness of the resulting BEV features at the target point by taking into account the local context around the reference point.}
    \label{persformer}
\end{figure}

\begin{figure}[b]
    \setlength{\abovecaptionskip}{0pt}
    \setlength{\belowcaptionskip}{0pt}
    \centering
    \includegraphics[width=1.0\linewidth]{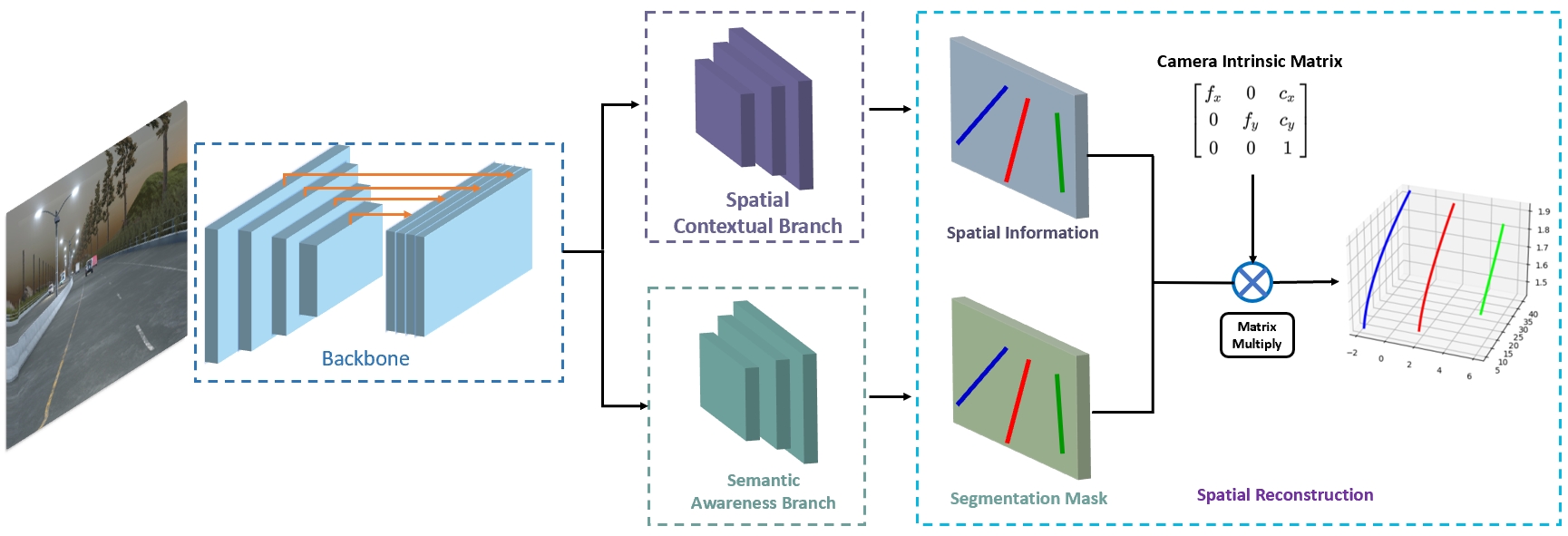}
    \captionsetup{font={footnotesize}}
    \caption{
    The architecture of \textcolor{black}{SALAD \cite{yan2022once}}. The input image is encoded into deep features by the backbone, and then two branches, the semantic awareness branch and the spatial contextual branch, decode the features to extract the spatial information of the lane and create the segmentation mask. Subsequently, the 3D reconstruction process integrates this information to ultimately derive the 3D lane positions in the real-world scene.}
    \label{ONCE_3DLanes}
\end{figure}

\textcolor{black}{The above methods are all CNN-based approaches.
PersFormer \cite{chen2022persformer} introduces the first Transformer-based approach for 3D lane detection and proposes a novel architecture called Perspective Transformer, as shown in Fig. \ref{persformer}. This Transformer-based architecture enables spatial feature transformation, allowing for accurate detection of 3D lane lines. Moreover, the proposed framework has the unique capability to simultaneously handle both 2D and 3D lane detection tasks, providing a unified solution. 
This method not only achieved the best performance at the time in the 3D lane detection task but also performes well in 2D lane detection. However, compared to previous methods, its inference time is longer, which slightly affects real-time performance.
Additionally, the work presents OpenLane, a large-scale 3D lane detection dataset built upon the influential Waymo Open dataset \cite{sun2020scalability}. OpenLane is the first dataset of its kind to offer high-quality annotations and diverse real-world scenarios, making it a valuable resource for advancing research in the field.}
\textcolor{black}{To avoid converting the feature map into a BEV image, Yan \textit{et al.} propose an extrinsic-free, anchor-free method called SALAD \cite{yan2022once}, as shown in Fig. \ref{ONCE_3DLanes}, which can directly regress the 3D information of lanes in the image view.
Specifically, the lane lines are segmented in the front view, and then the depth information is used to recover the 3D information of the lanes.
It suffers from certain limitations, since this method relies on LiDAR to provide 3D information for training, issues such as occlusion or point cloud sparsity at a distance can affect the algorithm's performance.
Additionally, this paper introducesd the largest real-world 3D lane detection dataset, ONCE-3DLanes dataset, and provides a more comprehensive evaluation metric to rekindle interest in this task in real-world scenarios.
}
\begin{figure}[h]
    \setlength{\abovecaptionskip}{0pt}
    \setlength{\belowcaptionskip}{0pt}
    \centering
    \includegraphics[width=1.0\linewidth]{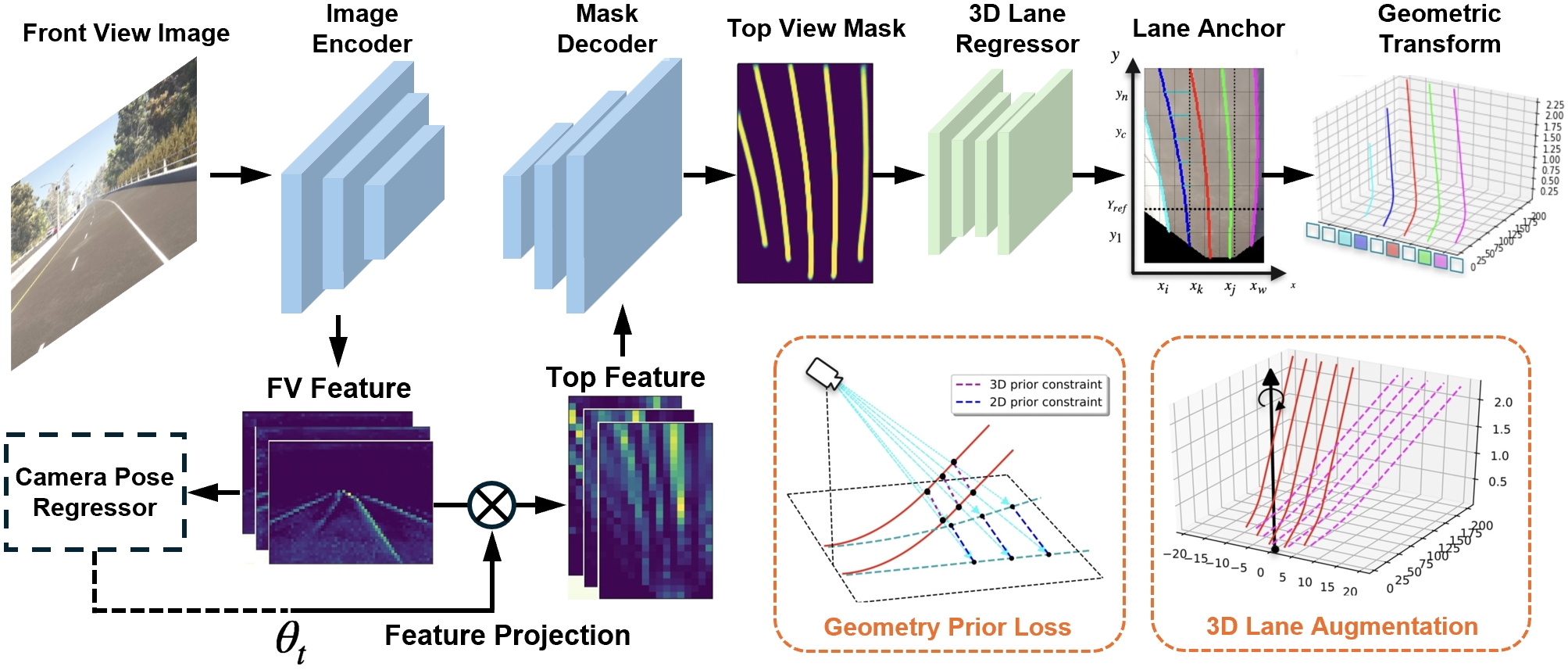}
    \captionsetup{font={footnotesize}}
    \caption{\textcolor{black}{The architecture of \cite{li2022reconstruct}}. The framework includes a top-view segmentation network and an anchor-based 3D lane regressor. Features from a front-view image are projected onto the top view to create the segmentation mask under top-view supervision. The anchor representation of 3D lanes is estimated from the top-view lane mask. 
    \textcolor{black}{Additionally, the geometry prior from 3D to 2D projection is incorporated for 3D lane reconstruction.}
}
    \label{reconstruct_from_bev}
\end{figure}
\textcolor{black}{Li \textit{et al.} \cite{li2022reconstruct} believe that the structure of 3D lanes and the 2D-3D relationships following geometric priors should be jointly optimized throughout the entire process of 3D lane detection.
They propose a novel loss function that utilizes the geometry structure prior of lanes in 3D space, enabling stable reconstruction from local to global and providing explicit supervision. It introduces a 2D lane feature extraction module that leverages direct supervision from the top view, ensuring maximum retention of lane structural information, especially in distant areas, the overall pipeline is shown in Fig. \ref{reconstruct_from_bev}. 
}
\textcolor{black}{
There are also two Transformer-based developments, CurveFormer \cite{bai2022curveformer} and CurveFormer++ \cite{bai2024curveformer++}, with CurveFormer++ being an improved version of CurveFormer. Their core idea is to represent lanes as a series of ordered 3D points, known as curve queries, and to iteratively optimize these queries in the Transformer decoder to accurately detect 3D lane lines. The improvements in CurveFormer++ involve incorporating temporal information fusion and a dynamic anchor set, as well as optimizing the network structure and cross-attention mechanism, achieving higher performance and better generalization capabilities in the 3D lane detection task. However, despite the progress made by CurveFormer and CurveFormer++ in the field of 3D lane detection, they also have certain drawbacks, such as high computational resource requirements and longer inference times.
}

\textcolor{black}{
Due to the challenge posed by different cameras with varying parameters in 3D lane detection, Wang \textit{et al.} introduce the concept of a Virtual Camera in BEV-LaneDet \cite{wang2023bev} to unify these parameters. In addition, the method proposes a simple yet effective 3D lane representation called Key-Points Representation, which is better suited for representing complex and diverse 3D lane structures. Moreover, a lightweight and chip-friendly Spatial Transformation Pyramid module was designed to convert multiscale front-view features into BEV features. This approach achieves excellent performance while maintaining fast inference speed. However, a drawback is that since this method focuses more on the BEV plane, its performance in terms of $z$-axis error is not ideal, requiring further improvement.}
Li \textit{et al.} propose a method \cite{li2022reconstruct} to directly extract top view lane information from front-view images, reducing the structural loss in the 2D lane representation. The overall pipeline of this method  is shown in the Fig. \ref{reconstruct_from_bev}.
In this work, the authors consider 3D lane detection as a reconstruction problem from 2D images to 3D space. They propose that explicit imposition of the geometric prior of 3D lanes during the training process is essential for fully utilizing the structural constraints of inter-lane and intra-lane relationships, and for extracting the height information of 3D lanes from the 2D lane representation. The authors analyze the geometric relationship between 3D lanes and their 2D representation, and propose an auxiliary loss function based on the geometric structure prior. They also demonstrate that explicit geometric supervision can enhance noise elimination, outlier rejection, and structure preservation for 3D lanes.
Similar to \cite{qin2020ultra}, Li \textit{et al.} propose GroupLane \cite{li2023grouplane}, a row-wise classification-based method for 3D lane detection.
GroupLane's design consists of two groups of convolutional heads, with each corresponding to a lane prediction. This grouping decouples the interaction of information among different lanes, reducing the difficulty in optimization. During training, predictions are matched with lane labels using a single-win one-to-one matching strategy, which assigns predictions to the most suitable labels for loss computation.
To address the challenge posed by the inevitable depth ambiguity in monocular images causing misalignment between the constructed surrogate feature map and the original image for lane detection, Luo \textit{et al.} propose a novel LATR model \cite{luo2023latr}. This is an end-to-end 3D lane detector that uses 3D-aware front-view features without the need for transformed view representation. Specifically, LATR detects 3D lanes through cross-attention based on query and key-value pairs constructed using a lane-aware query generator and dynamic 3D ground positional embedding. On one hand, each query is generated based on 2D lane-aware features and adopts a hybrid embedding to enhance lane information. On the other hand, 3D spatial information is injected as positional embedding from an iteratively updated 3D ground plane.
To address the issue of inaccurate view transformation caused by the neglect of road height changes in the process of transforming image-view features into BEV using IPM, Chen \textit{et al.} propose an efficient transformer for 3D lane detection \cite{chen2023efficient}. Unlike conventional transformers, the model includes a decomposed cross-attention mechanism that can simultaneously learn lane and BEV representations. This approach allows for more accurate view transformation compared to IPM-based methods and is more efficient.
Previous studies assume that all lanes are on a flat ground plane. However, Kim \textit{et al.} argue that algorithms based on this assumption have difficulty in detecting various lanes in actual driving environments and propose a new algorithm, D-3DLD \cite{kim2023d}. Unlike previous methods, this approach expands rich contextual features from the image domain to 3D space by utilizing depth-aware voxel mapping. Additionally, the method determines 3D lanes based on voxelized features. The authors design a new lane representation combined with uncertainties and predicted the confidence intervals of 3D lane points using Laplace loss.
Li \textit{et al.} propose a lightweight method \cite{li2022perspective} that uses MobileNet \cite{howard2017mobilenets} as the backbone network to reduce the demand for computational resources. The proposed method includes the following three stages. First, the MobileNet model is used to generate multi-scale front-view features from a single RGB image. Then, a perspective Transformer calculates the BEV features from the front-view features. Finally, two convolutional neural networks are used to predict the 2D and 3D coordinates and their respective lane types.
In paper \cite{han2023decoupling}, Han \textit{et al.} argue that the curve-based lane representation may not be well-suited for many irregular lane lines in real-world scenarios, which can lead to performance gaps compared to indirect representations such as segmentation-based or point-based methods. In this paper, the authors propose a new lane detection method, which can be decomposed into two parts: curve modeling and ground height regression. Specifically, a parameterized curve is used to represent lanes in the BEV space to reflect the original distribution of lanes. For the second part, since ground heights are determined by natural factors such as road conditions and they are not comprehensive, the ground heights are regressed separately from the curve modeling. Additionally, the authors have designed a new framework and a series of losses to unify the 2D and 3D lane detection tasks, guiding the optimization of models with or without 3D lane labels.

\textcolor{black}{The above methods only consider single-frame input, which leads to poor performance in scenarios without visual cues (e.g., obscured by surrounding vehicles). To address this issue and better exploit the spatio-temporal continuity of lanes, Wang \textit{et al.} propose STLane3D \cite{wang2022spatio}. This approach leverages a multi-frame fusion mechanism that utilizes the spatio-temporal continuity of lanes to resolve the problem of insufficient visual cues in single-frame detection. Additionally, they introduce a pre-alignment operation and a spatio-temporal attention module to effectively fuse multi-frame features, as well as a 3DLane IOULoss, which constrains the lateral and longitudinal changes of the predicted lanes to improve performance. The method achieves state-of-the-art results on public datasets.}

\textcolor{black}{
Learning-based methods cannot avoid relying on training data, however, annotating this training data is a high-cost and labor-intensive task, which hinders the application of 3D lane detection. In light of this, some researchers have proposed weakly supervised and self-supervised methods to reduce dependence on manually labeled data.
Ai \textit{et al.} propose a weakly supervised method called WS-3D-Lane \cite{ai2022ws}, which for the first time proposes a weakly supervised 3D lane detection method using only 2D lane labels, and outperform the previous 3D-LaneNet \cite{garnett20193d} method in evaluation. In addition, the authors propose a camera pitch self-calibration method, which can calculate the camera pitch angle online in real time, thereby reducing the error caused by the pitch angle change between the camera and the ground plane caused by the uneven road surface.
Although this method reduces the annotation requirements for 3D lane lines, it still relies on manually labeled 2D lane line data, and the assumption of constant lane width does not hold in some scenarios, which may affect the performance of the method.
To further reduce the reliance on manual data annotation, Nie \textit{et al.} propose LaneCorrect \cite{nie2024lanecorrect}, a self-supervised lane detection method. This approach introduces an unsupervised 3D lane segmentation technique that leverages the unique intensity of lanes in LiDAR point clouds to generate noisy 2D lane labels. Additionally, it presents a self-supervised training scheme that automatically corrects noisy lane labels by learning geometric consistency and instance-aware adversarial augmentation. Finally, the method introduces a distillation technique that allows the student network to be trained for any target lane detection dataset without using any labeled data.
Although this method significantly reduces the reliance on manual data annotation, the generated pseudo-labels contain some noises, such as projection errors and mislabeling caused by clustering errors, which can affect its performance. 
}


An intuitive summary of these methods is shown in Table \ref{tab:overviw}, which includes a description of the methods, the datasets used, the open-source status, and the network architecture.

\begin{table*}[h]
\captionsetup{font={footnotesize}}
\caption{ A detailed description of existing monocular 3D lane detection methods.} 
\centering
\begin{tabular}{m{0.4cm} m{0.4cm} m{7.0cm} m{1.8cm} m{0.5cm} m{1.4cm} m{3.1cm}}
\toprule
\footnotesize Ref. &\footnotesize Year &\footnotesize \makecell[c]{\footnotesize Description} & \footnotesize Dataset &\footnotesize Code  &\footnotesize \makecell[l]{Lane \\ Representation}  & \makecell[c]{\footnotesize Network\\ \hline \scriptsize Architecture \; Need BEV Trans.?} \\
\midrule
\scriptsize \cite{garnett20193d}  & \scriptsize 2018 & \scriptsize 
This work proposes a method that combines a novel network architecture, IPM projections, and anchor-based lane representation to achieve accurate 3D lane detection. & \scriptsize \makecell[l]{ Synthetic-3D-Lanes, \\ 3D-Lanes} &\scriptsize \makecell[l]{\hspace{0.0cm} N/A } &\textcolor{black}{\scriptsize \makecell[l]{Anchor \\ representation} }& \scriptsize \makecell[l]{\hspace{0.5cm} CNN    \hspace{1.2cm}   \ding{52} } \\
\hline 
\scriptsize \cite{guo2020gen}  & \scriptsize 2020 & \scriptsize 
This work employs a two-stage framework that decouples the learning of the image segmentation subnetwork from the geometry encoding subnetwork, thereby reducing the amount of 3D lane labels required for training.
& \scriptsize \makecell[l]{Apollo Synthetic.} & \scriptsize \makecell[l]{\hspace{0.0cm}
\textbf{\href{https://github.com/yuliangguo/Pytorch_Generalized_3D_Lane_Detection}{link}} } & \textcolor{black}{\scriptsize \makecell[l]{Anchor \\ representation}} & \scriptsize \makecell[l]{\hspace{0.5cm} CNN    \hspace{1.2cm}   \ding{52} }\\
\hline 
\scriptsize \cite{efrat2020semi} &\scriptsize 2020 &\scriptsize 
This approach introduces uncertainty estimation, utilizes a semi-local BEV tile representation to decompose lanes into simple segments, and subsequently clusters these segments into complete lanes.
&\scriptsize \makecell[l]{Synthetic-3D-Lanes,\\3D-lanes.} &\scriptsize \makecell[l]{\hspace{0.0cm} N/A  } & \textcolor{black}{\scriptsize \makecell[l]{Tile \\ representation}} &\scriptsize \makecell[l]{\hspace{0.5cm} CNN    \hspace{1.2cm}   \ding{52} }\\
\hline 
\scriptsize \cite{efrat20203d} &\scriptsize 2020 &\scriptsize 
This method is anchor-free and capable of detecting complex lane topologies, including splits, merges, short lanes, and perpendicular lanes.
&\scriptsize \makecell[l]{Synthetic-3D-Lanes,\\Real-3D-lanes} &\scriptsize \makecell[l]{\hspace{0.0cm} N/A } & \textcolor{black}{\scriptsize \makecell[l]{Tile \\ representation}} &\scriptsize \makecell[l]{\hspace{0.5cm} CNN    \hspace{1.2cm}   \ding{52} }\\
\hline
\scriptsize \cite{liu2022learning} &\scriptsize 2021 &\scriptsize 
This work presents a two-stage framework that first leverages auxiliary 3D lane tasks and geometry constraints, and then performs accurate 3D lane prediction in the second stage using the estimated pose.
&\scriptsize \makecell[l]{Apollo Synthetic.} &\scriptsize \makecell[l]{\hspace{0.0cm} \textbf{\href{https://github.com/liuruijin17/CLGo}{link}} } & \textcolor{black}{\scriptsize \makecell[l]{Polynomial \\ representation} } &\scriptsize \makecell[l]{\hspace{0.4cm} \textcolor{black}{Hybrid}    \hspace{1.1cm}   \ding{52} }\\
\hline
\scriptsize \cite{chen2022persformer}&\scriptsize 2022 &\scriptsize This paper introduces a novel Transformer-based spatial feature transformation module and presents OpenLane, a large-scale real-world 3D lane dataset. &\scriptsize \makecell[l]{Apollo Synthetic,\\ OpenLane.} &\scriptsize \makecell[l]{\hspace{0.0cm} \textbf{\href{https://github.com/OpenDriveLab/PersFormer_3DLane}{link}}} & \textcolor{black}{\scriptsize \makecell[l]{Anchor \\ representation}} &\scriptsize \makecell[l]{\hspace{0.3cm} Transformer    \hspace{0.7cm}   \ding{52} }\\ 
\hline
\scriptsize \cite{yan2022once}&\scriptsize 2022&\scriptsize This paper introduces the ONCE-3DLanes dataset and proposes SALAD, an extrinsic-free, anchor-free method for regressing 3D lane coordinates in image view. &\scriptsize \makecell[l]{ONCE-3DLanes.} 
&\scriptsize \makecell[l]{\hspace{0.0cm} \textbf{\href{https://github.com/once-3dlanes/once_3dlanes_benchmark}{link}}} & \textcolor{black}{\scriptsize \makecell[l]{Pixel \\ representation} } &\scriptsize \makecell[l]{ \hspace{0.5cm} CNN    \hspace{1.2cm}   \ding{56} }\\  
\hline
\scriptsize \cite{jiang2022robust} &\scriptsize 2022 &
\scriptsize This paper proposes Att-Gen-LaneNet, a two-stage network for robust 3D lane detection. It integrates the Efficient Channel Attention (ECA) and Convolutional Block Attention Module (CBAM) to improve segmentation and generalization. &\scriptsize \makecell[l]{Apollo Synthetic.} &\scriptsize \makecell[l]{\hspace{0.0cm} N/A } & \textcolor{black}{\scriptsize \makecell[l]{Anchor \\ representation}} &\scriptsize \makecell[l]{ \hspace{0.5cm} CNN    \hspace{1.2cm}   \ding{52}}\\
\hline
\scriptsize \cite{li2022reconstruct} &\scriptsize 2022 &\scriptsize This paper proposes an approach for monocular 3D lane detection by leveraging the geometry structure and explicit supervision based on the structure prior.  &\scriptsize \makecell[l]{Apollo Synthetic.} &\scriptsize \makecell[l]{\hspace{0.0cm} N/A } & \textcolor{black}{\scriptsize \makecell[l]{Anchor \\ representation}} &\scriptsize \makecell[l]{ \hspace{0.5cm} CNN    \hspace{1.2cm}   \ding{52}}\\
\hline
\scriptsize \cite{li2022perspective} &\scriptsize 2022 &\scriptsize 
This method comprises three stages: first, generating multiscale front-view features; second, calculating BEV features; and third, predicting both 2D and 3D lanes.
&\scriptsize \makecell[l]{ OpenLane.} &\scriptsize \makecell[l]{\hspace{0.0cm} N/A }& \textcolor{black}{\scriptsize \makecell[l]{Anchor \\ representation}}  &\scriptsize \makecell[l]{ \hspace{0.5cm} CNN    \hspace{1.2cm}   \ding{52} } \\
\hline
\scriptsize \cite{bai2022curveformer} &\scriptsize 2022 &\scriptsize This method formulates 3D lane detection as a curve propagation problem, enhancing performance by leveraging curve queries, a curve cross-attention module, and a context sampling module to capture relevant image features. &\scriptsize \makecell[l]{Apollo Synthetic,\\ OpenLane.} &\scriptsize \makecell[l]{\hspace{0.0cm} N/A } & \textcolor{black}{\scriptsize \makecell[l]{Anchor \\ representation}}  &\scriptsize \makecell[l]{\hspace{0.3cm} Transformer    \hspace{0.7cm}   \ding{56} }\\
\hline
\scriptsize \cite{ai2022ws} &\scriptsize 2022&\scriptsize This paper introduces a method training 3D lanes with only 2D lane labels, leveraging assumptions of constant lane width and equal height on adjacent lanes for indirect supervision of 3D lane heights. &\scriptsize \makecell[l]{Apollo Synthetic,\\ONCE-3DLanes.} &\scriptsize \makecell[l]{\hspace{0.0cm} N/A} & \textcolor{black}{\scriptsize \makecell[l]{Anchor \\ representation}} &\scriptsize \makecell[l]{ \hspace{0.5cm} CNN    \hspace{1.2cm}   \ding{52} } \\
\hline
\scriptsize \cite{wang2023bev}&\scriptsize 2022 &\scriptsize
This paper introduces a Virtual Camera for spatial relationship consistency, a Key-Points Representation module for diverse 3D lane structures, and a Spatial Transformation Pyramid for front-view to BEV feature transformation. &\scriptsize \makecell[l]{Apollo Synthetic,\\ OpenLane.} &\scriptsize \makecell[l]{\hspace{0.0cm} \textbf{\href{https://github.com/gigo-team/bev_lane_det}{link}}} & \textcolor{black}{\scriptsize \makecell[l]{Key-points \\ representation}}  &\scriptsize \makecell[l]{ \hspace{0.5cm} CNN    \hspace{1.2cm}   \ding{52} }\\
\hline
\scriptsize \cite{wang2022spatio}&\scriptsize 2022 &\scriptsize 
This model utilizes multi-frame input and removes dependency on camera pose by integrating a multi-frame pre-alignment layer, a spatio-temporal attention module, and 3DLane IOULoss.
&\scriptsize \makecell[l]{OpenLane,\\ ONCE-3DLanes.} &\scriptsize \makecell[l]{\hspace{0.0cm} N/A} & \textcolor{black}{\scriptsize \makecell[l]{Anchor \\ representation}} &\scriptsize \makecell[l]{ \hspace{0.5cm} CNN     \hspace{1.2cm}   \ding{52} }\\ 
\hline
\scriptsize \cite{huang2023anchor3dlane} &\scriptsize 2023 &\scriptsize
This paper introduces Anchor3DLane, a BEV-free method predicting 3D lanes directly from front-viewed (FV) representations using defined 3D lane anchors. &\scriptsize \makecell[l]{Apollo Synthetic,\\ OpenLane,\\ONCE-3DLanes.} &\scriptsize \makecell[l]{\hspace{0.0cm} \textbf{\href{https://github.com/tusen-ai/Anchor3DLane}{link}}} & \textcolor{black}{\scriptsize \makecell[l]{Anchor \\ representation}}  &\scriptsize \makecell[l]{ \hspace{0.4cm} \textcolor{black}{Hybrid}    \hspace{1.12cm}   \ding{56} }\\
\hline
\scriptsize \cite{kim2023d} &\scriptsize 2023 &\scriptsize
This paper introduces a method that employs depth-aware voxel mapping, extends contextual features to 3D space, and predicts confidence intervals using Laplace loss. &\scriptsize \makecell[l]{LLAMAS-3DLD,\\ Pandaset-3DLD,\\Apollo Synthetic.} &\scriptsize \makecell[l]{\hspace{0.0cm} N/A}  & \textcolor{black}{\scriptsize \makecell[l]{Anchor \\ representation}} &\scriptsize \makecell[l]{ \hspace{0.5cm} CNN   \hspace{1.2cm}   \ding{52} } \\
\hline
\scriptsize \cite{chen2023efficient} &\scriptsize 2023 &\scriptsize
This paper proposes an efficient transformer for 3D lane detection using a decomposed cross-attention mechanism to learn lane and BEV representations concurrently. &\scriptsize \makecell[l]{OpenLane,\\ ONCE-3DLanes.}  &\scriptsize \makecell[l]{\hspace{0.0cm} N/A } & \textcolor{black}{\scriptsize \makecell[l]{Key-points \\ representation}} &\scriptsize \makecell[l]{\hspace{0.3cm}  Transformer    \hspace{0.7cm}   \ding{56} }\\
\hline
\scriptsize \cite{li2023grouplane} &\scriptsize 2023 &\scriptsize 
This work, similar to \cite{qin2020ultra}, pioneers 3D lane detection by utilizing fully convolutional heads based on row-wise classification.
&\scriptsize \makecell[l]{OpenLane,\\ ONCE-3DLanes,\\OpenLane-Huawei.} &\scriptsize \makecell[l]{\hspace{0.0cm} N/A}  & \textcolor{black}{\scriptsize \makecell[l]{Row-wise \\ representation}} &\scriptsize \makecell[l]{ \hspace{0.5cm} CNN    \hspace{1.2cm}   \ding{52} }\\
\hline
\scriptsize \cite{luo2023latr} &\scriptsize 2023 &\scriptsize
This paper introduces LATR, an end-to-end 3D lane detection model utilizing 3D-aware front-view features and cross-attention mechanism with lane-aware query generator. &\scriptsize \makecell[l]{OpenLane,\\ Apollo Synthetic,\\ONCE-3DLanes.}  &\scriptsize \makecell[l]{\hspace{0.0cm} \textbf{\href{https://github.com/JMoonr/LATR}{link}}} & \textcolor{black}{\scriptsize \makecell[l]{Key-points  \\ representation}} &\scriptsize \makecell[l]{ \hspace{0.3cm}Transformer    \hspace{0.8cm}   \ding{56} }\\
\hline
\scriptsize \cite{han2023decoupling}&\scriptsize 2023 &\scriptsize
This paper introduces a novel lane detection method, dividing it into curve modeling and ground height regression, within a unified framework for both 2D and 3D lane detection.&\scriptsize \makecell[l]{OpenLane,\\ ONCE-3DLanes, \\ TuSimple, CULane.} &\scriptsize \makecell[l]{\hspace{0.0cm} N/A} & \textcolor{black}{\scriptsize \makecell[l]{Anchor \\ representation}} &\scriptsize \makecell[l]{ \hspace{0.5cm} CNN    \hspace{1.2cm}   \ding{56} }\\ 
\hline
\scriptsize \cite{bai2024curveformer++}&\scriptsize 2024 &\scriptsize
This approach models the 3D detection task as a curve propagation problem, where each lane is represented by a curve query with a dynamic and ordered anchor point set. &\scriptsize \makecell[l]{OpenLane, \\ ONCE-3DLanes.}  &\scriptsize \makecell[l]{\hspace{0.0cm} N/A} & \textcolor{black}{\scriptsize \makecell[l]{Anchor \\ representation}} &\scriptsize \makecell[l]{\hspace{0.3cm}Transformer    \hspace{0.8cm}   \ding{56} }\\ 
\hline
\scriptsize \textcolor{black}{\cite{nie2024lanecorrect}} &\scriptsize \textcolor{black}{2024} &\scriptsize \textcolor{black}{
This paper introduces a self-supervised lane detection method that uses LiDAR intensity to generate noisy lane labels. Through geometric consistency and adversarially enhanced instance awareness, it corrects these labels without requiring manually annotated data.} &\scriptsize \makecell[l]{\textcolor{black}{OpenLane,} \\ \textcolor{black}{ONCE-3DLanes.}}  &\scriptsize \makecell[l]{\hspace{0.0cm} \textcolor{black}{N/A}} & \textcolor{black}{\scriptsize \makecell[l]{Anchor \\ representation}} &\scriptsize \makecell[l]{\hspace{0.5cm} \textcolor{black}{CNN}  \hspace{1.2cm}   \ding{56} }\\
\bottomrule
\end{tabular}
\label{tab:overviw}
\end{table*}


\begin{table}[h]
\caption{LIST OF USED VARIABLES}
\label{tab:notations}
\centering
\begin{tabular}{c c|c c c }
\cline{1-4}
\multicolumn{2}{c|}{\makecell[c]{Notation}} & \multicolumn{2}{c}{ \makecell[c]{Description}} & \\
\cline{1-4}
\multicolumn{2}{c|}{\multirow{2}{*}{\makecell[c]{\scriptsize $Accuracy$ }}} & \multicolumn{2}{l}{ \multirow{2}{*}{\makecell[l]{\scriptsize The proportion of correctly classified samples to the total \\ number of samples.}}} \\
 & & \\
\cline{1-4}
\multicolumn{2}{c|}{\multirow{2}{*}{\makecell[c]{\scriptsize $Recall$ }}} & \multicolumn{2}{l}{ \multirow{2}{*}{\scriptsize \makecell[l]{ The proportion of samples classified as positive and predicted\\ correctly to the total number of true positive samples. }}} \\
  & & \\
\cline{1-4}
\multicolumn{2}{c|}{\multirow{3}{*}{\makecell[c]{\scriptsize $Precision$ }}} & \multicolumn{3}{l}{ \multirow{3}{*}{\scriptsize \makecell[l]{The proportion of samples classified as positive and predicted \\ correctly to the total number of samples classified as \\ positive. }}} \\
& & \\
& & \\
\cline{1-4}
\multicolumn{2}{c|}{\multirow{1}{*}{\makecell[c]{\scriptsize $F-Score$ }}} & \multicolumn{2}{l}{ \multirow{1}{*}{\makecell[l]{\scriptsize The harmonic mean of precision and recall. }}} \\
\cline{1-4}
\multicolumn{2}{c|}{\multirow{2}{*}{\makecell[c]{\scriptsize $\beta$ }}} & \multicolumn{2}{l}{ \multirow{2}{*}{\scriptsize \makecell[l]{The parameter used to measure the importance of precision \\ and recall in F-Score. }}} \\
    & & \\
\cline{1-4}
\multicolumn{2}{c|}{\multirow{1}{*}{\makecell[c]{\scriptsize $AP$ }}} & \multicolumn{2}{l}{ \multirow{1}{*}{\makecell[l]{\scriptsize Average Precision. }}} \\
\cline{1-4}
\multicolumn{2}{c|}{\multirow{1}{*}{\makecell[c]{\scriptsize $CD$ }}} & \multicolumn{2}{l}{ \multirow{1}{*}{\makecell[l]{\scriptsize Chamfer distance. }}} \\
\cline{1-4}
\multicolumn{2}{c|}{\multirow{1}{*}{\makecell[c]{\scriptsize \textcolor{black}{ $P$} }}} & \multicolumn{2}{l}{ \multirow{1}{*}{\makecell[l]{\scriptsize \textcolor{black}{ A set of prediction lane points.} }}} \\
\cline{1-4}
\multicolumn{2}{c|}{\multirow{1}{*}{\makecell[c]{\scriptsize \textcolor{black}{ $Q$ }}}} & \multicolumn{2}{l}{ \multirow{1}{*}{\makecell[l]{\scriptsize \textcolor{black}{ A set of ground truth lane points.} }}} \\
\cline{1-4}
\multicolumn{2}{c|}{\multirow{1}{*}{\makecell[c]{\scriptsize \textcolor{black}{ $p$ }}}} & \multicolumn{2}{l}{ \multirow{1}{*}{\makecell[l]{\scriptsize \textcolor{black}{A point of $ P $.} }}} \\
\cline{1-4}
\multicolumn{2}{c|}{\multirow{1}{*}{\makecell[c]{\scriptsize \textcolor{black}{ $q$ }}}} & \multicolumn{2}{l}{ \multirow{1}{*}{\makecell[l]{\scriptsize \textcolor{black}{ A point of $Q$.} }}} \\
\cline{1-4}
\multicolumn{2}{c|}{\multirow{3}{*}{\makecell[c]{\scriptsize $IoU$ }}} & \multicolumn{3}{l}{ \multirow{3}{*}{\scriptsize \makecell[l]{Intersection over Union (IoU), used to measure the degree of \\ overlap between predicted bounding boxes and ground truth \\ bounding boxes. }}} \\
    & & \\
    & & \\
\cline{1-4}
\multicolumn{2}{c|}{\multirow{1}{*}{\makecell[c]{\scriptsize $mIoU$ }}} & \multicolumn{2}{l}{ \multirow{1}{*}{\makecell[l]{\scriptsize The average value of IoU. }}} \\
\cline{1-4}
\multicolumn{2}{c|}{\multirow{1}{*}{\makecell[c]{\scriptsize $MSE$ }}} & \multicolumn{2}{l}{ \multirow{1}{*}{\makecell[l]{\scriptsize Mean Square Error. }}} \\
\cline{1-4}
\multicolumn{2}{c|}{\multirow{1}{*}{\makecell[c]{\scriptsize $MAE$ }}} & \multicolumn{2}{l}{ \multirow{1}{*}{\makecell[l]{\scriptsize Mean Absolute Error. }}} \\
\cline{1-4}
\multicolumn{2}{c|}{\multirow{2}{*}{\makecell[c]{\scriptsize $CE$ }}} & \multicolumn{2}{l}{ \multirow{2}{*}{\makecell[l]{\scriptsize Cross-Entropy, a commonly used loss function for evaluating the \\ performance of classification models.}}} \\
& & \\
\cline{1-4}
\multicolumn{2}{c|}{\multirow{2}{*}{\makecell[c]{\scriptsize $BCE$ }}} & \multicolumn{2}{l}{ \multirow{2}{*}{\makecell[l]{\scriptsize Binary Cross-Entropy, a common loss function used in binary \\ classification tasks.}}} \\
    & & \\
\cline{1-4}
\multicolumn{2}{c|}{\multirow{1}{*}{\makecell[c]{\scriptsize $y$ }}} & \multicolumn{2}{l}{ \multirow{1}{*}{\makecell[l]{\scriptsize Ground truth value. }}} \\
\cline{1-4}
\multicolumn{2}{c|}{\multirow{1}{*}{\makecell[c]{\scriptsize $\hat{y}$ }}} & \multicolumn{2}{l}{ \multirow{1}{*}{\makecell[l]{\scriptsize Network's predicted value. }}} \\
\cline{1-4}
\multicolumn{2}{c|}{\multirow{1}{*}{\makecell[c]{\scriptsize $p_i$ }}} & \multicolumn{2}{l}{ \multirow{1}{*}{\makecell[l]{\scriptsize The probability of the model predicting the positive class. }}} \\
\cline{1-4}
\multicolumn{2}{c|}{\multirow{1}{*}{\makecell[c]{\scriptsize $p_t$ }}} & \multicolumn{2}{l}{ \multirow{1}{*}{\makecell[l]{\scriptsize The probability predicted for the actual class, used in Huber loss.}}} \\
\cline{1-4}
\multicolumn{2}{c|}{\multirow{1}{*}{\makecell[c]{\scriptsize $p_{ic}$ }}} & \multicolumn{2}{l}{ \multirow{1}{*}{\makecell[l]{\scriptsize The predicted probability of sample $i$ belonging to class $c$. }}} \\
\cline{1-4}
\multicolumn{2}{c|}{\multirow{1}{*}{\makecell[c]{\scriptsize $y_{ic}$ }}} & \multicolumn{2}{l}{ \multirow{1}{*}{\makecell[l]{\scriptsize The ground true label of sample $i$ for class $c$. }}} \\
\cline{1-4}
\multicolumn{2}{c|}{\multirow{1}{*}{\makecell[c]{\scriptsize $log$ }}} & \multicolumn{2}{l}{ \multirow{1}{*}{\makecell[l]{\scriptsize Logarithmic function. }}} \\
\cline{1-4}
\multicolumn{2}{c|}{\multirow{2}{*}{\makecell[c]{\scriptsize $\gamma$ }}} & \multicolumn{2}{l}{ \multirow{2}{*}{\makecell[l]{\scriptsize Threshold parameter in Huber loss, defining the point where \\ the loss function transitions from quadratic to linear.}}} \\
    & & \\
\cline{1-4}
\multicolumn{2}{c|}{\multirow{1}{*}{\makecell[c]{\scriptsize $| \cdot |$ }}} & \multicolumn{2}{l}{ \multirow{1}{*}{\makecell[l]{\scriptsize The size of the set (usually the number of pixels). }}} \\
\cline{1-4}
\multicolumn{2}{c|}{\multirow{1}{*}{\makecell[c]{\scriptsize $\cup$ }}} & \multicolumn{2}{l}{ \multirow{1}{*}{\makecell[l]{\scriptsize Union operation. }}} \\
\cline{1-4}
\multicolumn{2}{c|}{\multirow{1}{*}{\makecell[c]{\scriptsize $\cap$ }}} & \multicolumn{2}{l}{ \multirow{1}{*}{\makecell[l]{\scriptsize  Intersection operation.}}} \\
\cline{1-4}
\multicolumn{2}{c|}{\multirow{1}{*}{\makecell[c]{\scriptsize $S$ }}} & \multicolumn{2}{l}{ \multirow{1}{*}{\makecell[l]{\scriptsize The ground truth region, used in IoU loss. }}} \\
\cline{1-4}
\multicolumn{2}{c|}{\multirow{1}{*}{\makecell[c]{\scriptsize $\hat{S}$ }}} & \multicolumn{2}{l}{ \multirow{1}{*}{\makecell[l]{\scriptsize The region predicted by the network, used in IoU loss. }}} \\
\cline{1-4}
\multicolumn{2}{c|}{\multirow{1}{*}{\makecell[c]{\scriptsize $N $ }}} & \multicolumn{2}{l}{ \multirow{1}{*}{\makecell[l]{\scriptsize The sample size. }}} \\
\cline{1-4}
\multicolumn{2}{c|}{\multirow{1}{*}{\makecell[c]{\scriptsize $C$ }}} & \multicolumn{2}{l}{ \multirow{1}{*}{\makecell[l]{\scriptsize The number of categories. }}} \\

\cline{1-4}

\end{tabular}
\end{table}

\begin{table*}[h!]
\scriptsize
\centering
\begin{tabular}{c|c|c c c c c c c c}
\hline
\textbf{Scene} & \textbf{Method} & \textbf{FPS}\textuparrow & \textbf{Device} & \textbf{AP(\%)}\textuparrow & \textbf{F1(\%)}\textuparrow & \textbf{x err/N(m)}\textdownarrow & \textbf{$x$ err/F(m)}\textdownarrow & \textbf{$z$ err/N(m)}\textdownarrow & \textbf{$z$ err/F(m)}\textdownarrow \\ 
\hline
\multirow{9}{*}{\makecell[c]{Balanced \\ Scene} }
& 3DLaneNet  \cite{garnett20193d} & \textcolor{black}{91} & \textcolor{black}{RTX 3080Ti} & 89.3 & 86.4 & 0.068 & 0.477 & 0.015 & \textbf{0.202} \\ 
 & Gen-LaneNet \cite{guo2020gen} & \textcolor{black}{102} & \textcolor{black}{RTX 3080Ti} & 90.1 & 88.1 & 0.061 & 0.496 & 0.012 & 0.214 \\ 
 & CLGo  \cite{liu2022learning} & \textcolor{black}{128} & \textcolor{black}{RTX 3080Ti} & 94.2 & 91.9 & 0.061 & 0.361 & 0.029 & 0.250 \\ 
 & PersFormer  \cite{chen2022persformer} & \textcolor{black}{29} & \textcolor{black}{RTX 3080Ti} & - & 92.9 & 0.054 & 0.356 & 0.010 & 0.234 \\ 
 & Reconstruct from Top  \cite{li2022reconstruct} & 82 & GTX 1080Ti & 93.8 & 91.9 & 0.049 & 0.387 & 0.008 & 0.213 \\ 
 & CurveFormer  \cite{bai2022curveformer} & - & - & 97.3 & 95.8 & 0.078 & 0.326 & 0.018 & 0.219 \\ 
 & WS-3D-Lane  \cite{ai2022ws} & 26.7 & RTX 3080Ti & 95.7 & 93.5 & 0.027 & 0.321 & \textbf{0.006} & 0.215 \\ 
 & Anchor3DLane  \cite{huang2023anchor3dlane} & \textcolor{black}{148} & \textcolor{black}{RTX 3080Ti} & 97.2 & 95.6 & 0.052 & 0.306 & 0.015 & 0.223 \\ 
 & BEV-LaneDet  \cite{wang2023bev} & \textcolor{black}{158} & \textcolor{black}{RTX 3080Ti} & - & \textbf{98.7} & \textbf{0.016} & \textbf{0.242} & 0.020 & 0.216 \\ 
& LATR \cite{luo2023latr} & \textcolor{black}{14} & \textcolor{black}{RTX 3080Ti} & \textbf{97.9} & 96.8 & 0.022 & 0.253 & 0.007 & \textbf{0.202} \\ 
& D-3DLD \cite{kim2023d} & 22.6 & Tesla A100 & 95.5 & 93.3 & 0.039 & 0.412 & 0.011 & 0.234 \\
\hline
\multirow{9}{*}{\makecell[c]{Rarely \\ Observed}} 
& 3DLaneNet  \cite{garnett20193d} & \textcolor{black}{91} & \textcolor{black}{RTX 3080Ti} & 74.6 & 72.0 & 0.166 & 0.855 & 0.039 & \textbf{0.521} \\ 
 & Gen-LaneNet  \cite{guo2020gen} & \textcolor{black}{102} & \textcolor{black}{RTX 3080Ti} & 79.0 & 78.0 & 0.139 & 0.903 & 0.030 & 0.539 \\ 
 & CLGo  \cite{liu2022learning} & \textcolor{black}{128} & \textcolor{black}{RTX 3080Ti} & 88.3 & 86.1 & 0.147 & 0.735 & 0.071 & 0.609 \\ 
 & PersFormer  \cite{chen2022persformer} & \textcolor{black}{29} & \textcolor{black}{RTX 3080Ti} & - & 87.5 & 0.107 & 0.782 & 0.024 & 0.602 \\ 
 & Reconstruct from Top  \cite{li2022reconstruct} & 82 & GTX 1080Ti & 85.2 & 83.7 & 0.126 & 0.903 & 0.023 & 0.625 \\ 
 & CurveFormer  \cite{bai2022curveformer} & - & - & 97.1 & 95.6 & 0.182 & 0.737 & 0.039 & 0.561 \\ 
 & WS-3D-Lane  \cite{ai2022ws} & 26.7 & RTX 3080Ti & 88.0 & 86.0 & 0.070 & 0.741 & \textbf{0.015} & 0.585 \\ 
 & Anchor3DLane  \cite{huang2023anchor3dlane} & \textcolor{black}{148} & \textcolor{black}{RTX 3080Ti} & 96.9 & 94.4 & 0.094 & 0.693 & 0.027 & 0.579 \\ 
 & BEV-LaneDet  \cite{wang2023bev} & \textcolor{black}{158} & \textcolor{black}{RTX 3080Ti} & - & \textbf{99.1} & \textbf{0.031} & \textbf{0.594} & 0.040 & 0.556 \\ 
 & LATR  \cite{luo2023latr} & \textcolor{black}{14} & \textcolor{black}{RTX 3080Ti} & \textbf{97.3} & 96.1 & 0.050 & 0.600 & \textbf{0.015} & 0.532 \\ 
\hline
\multirow{9}{*}{\makecell[c]{Visual \\ Variations}} 
& 3D-LaneNet  \cite{garnett20193d} & \textcolor{black}{91} & \textcolor{black}{RTX 3080Ti} & 74.9 & 72.5 & 0.115 & 0.601 & 0.032 & 0.230 \\ 
 & Gen-LaneNet  \cite{guo2020gen} & \textcolor{black}{102} & \textcolor{black}{RTX 3080Ti} & 87.2 & 85.3 & 0.074 & 0.538 & 0.015 & 0.232 \\ 
 & CLGo \cite{liu2022learning} & \textcolor{black}{128} & \textcolor{black}{RTX 3080Ti} & 89.2 & 87.3 & 0.084 & 0.464 & 0.045 & 0.312 \\ 
 & PersFormer  \cite{chen2022persformer} & \textcolor{black}{29} & \textcolor{black}{RTX 3080Ti} & - & 89.6 & 0.074 & 0.430 & 0.015 & 0.266 \\ 
 & Reconstruct from Top \cite{li2022reconstruct} & 82 & GTX 1080Ti & 92.1 & 89.9 & 0.060 & 0.446 & 0.011 & 0.235 \\ 
 & CurveFormer \cite{bai2022curveformer} & - & - & 93.0 & 90.8 & 0.125 & 0.410 & 0.028 & 0.254 \\ 
 & WS-3D-Lane  \cite{ai2022ws} & 26.7 & RTX 3080Ti & 95.8 & 93.6 & \textbf{0.025} & 0.338 & \textbf{0.005} & 0.237 \\ 
 & Anchor3DLane  \cite{huang2023anchor3dlane} & \textcolor{black}{148} & \textcolor{black}{RTX 3080Ti} & 93.6 & 91.4 & 0.068 & 0.367 & 0.020 & 0.232 \\ 
 & BEV-LaneDet \cite{wang2023bev} & \textcolor{black}{158} & \textcolor{black}{RTX 3080Ti} & - & \textbf{96.9} & 0.027 & 0.320 & 0.031 & 0.256 \\ 
 & LATR  \cite{luo2023latr} & \textcolor{black}{14} & \textcolor{black}{RTX 3080Ti} & \textbf{96.6} & 95.1 & 0.045 & \textbf{0.315} & 0.016 & \textbf{0.228} \\ 
\hline
\end{tabular}
\caption{An in-depth comparison of various methods on the ApolloSim dataset under three different split settings. “N” and “F” are short for near and far, respectively.}
\label{performance_comparison}
\end{table*}

\begin{table*}[h!]
\scriptsize
\centering
\begin{tabular}{c|c c c c c c c c c }
\hline
\textbf{Method} & \textbf{FPS}\textuparrow & \textbf{Device} & \textbf{All} & \textbf{Up \& Down} & \textbf{Curve} & \textbf{Extreme Weather} & \textbf{Night} & \textbf{Intersection} & \textbf{Merge \& Split} \\ 
\hline
3D-LaneNet \cite{garnett20193d} & \textcolor{black}{91} & \textcolor{black}{RTX 3080Ti} & 44.1 & 40.8 & 46.5 & 47.5 & 41.5 & 32.1 & 41.7 \\ 
GenLaneNet \cite{guo2020gen} & \textcolor{black}{102} & \textcolor{black}{RTX 3080Ti} & 32.3 & 25.4 & 33.5 & 28.1 & 18.7 & 21.4 & 31.0 \\ 
PersFormer  \cite{chen2022persformer} & \textcolor{black}{29} & \textcolor{black}{RTX 3080Ti} & 50.5 & 42.4 & 55.6 & 48.6 & 46.6 & 40.0 & 50.7 \\ 
CurveFormer \cite{bai2022curveformer} & - & - & 50.5 & 45.2 & 56.6 & 49.7 & 49.1 & 42.9 & 45.4 \\ 
Anchor3DLane  \cite{huang2023anchor3dlane} & \textcolor{black}{148} & \textcolor{black}{RTX 3080Ti} & 54.3 & 47.2 & 58.0 & 52.7 & 48.7 & 45.8 & 51.7 \\ 
BEV-LaneDet  \cite{wang2023bev} & \textcolor{black}{158} & \textcolor{black}{RTX 3080Ti} & 58.4 & 48.7 & 63.1 & 53.4 & 53.4 & 50.3 & 53.7 \\ 
Efficient Transformer \cite{chen2023efficient} & 72 & Tesla V100 & 63.8 & \textbf{57.6} & \textbf{73.2} & \textbf{57.3} & \textbf{59.7} & \textbf{57.0} & \textbf{63.9} \\ 
GroupLane  \cite{li2023grouplane} & 38.2 & RTX 2080 & \textbf{64.1} & - & - & - & - & - & - \\ 
LATR  \cite{luo2023latr} & \textcolor{black}{14} & \textcolor{black}{RTX 3080Ti} & 61.9 & 55.2 & 68.2 & 57.1 & 55.4 & 52.3 & 61.5 \\
DecoupleLane  \cite{han2023decoupling} & - & - & 51.2 & 43.5 & 57.3 & - & 48.9 & 43.5 & - \\
CurveFormer++ \cite{bai2024curveformer++} & 11.4 & RTX 3090 & 52.7 & 48.3 & 59.4 & 50.6 & 48.4 & 45.0 & 48.1 \\
\hline
\end{tabular}
\caption{Comparison of various 3D methods on the OpenLane benchmark, evaluating the F-Score on both the entire validation set and separate sets corresponding to five different scenarios.}
\label{table2}
\end{table*}

\begin{table*}[h!]
\centering
\begin{tabular}{c|c c c c c c }
\hline
\textbf{Method} & \textbf{FPS}\textuparrow & \textbf{Device} & \textbf{F1(\%)}\textuparrow & \textbf{Precision(\%)}\textuparrow & \textbf{Recall(\%)}\textuparrow & \textbf{CD Error(m)}\textdownarrow \\ 
\hline
3D-LaneNet  \cite{garnett20193d} & \textcolor{black}{91} & \textcolor{black}{RTX 3080Ti} & 44.73 & 61.46 & 35.16 & 0.127 \\ 
Gen-LaneNet \cite{guo2020gen} & \textcolor{black}{102} & \textcolor{black}{RTX 3080Ti} & 45.59 & 63.95 & 35.42 & 0.121 \\ 
SALAD  \cite{yan2022once} & - & - & 64.07 & 75.90 & 55.42 & 0.098 \\ 
PersFormer  \cite{chen2022persformer} & \textcolor{black}{29} & \textcolor{black}{RTX 3080Ti} & 74.33 & 80.30 & 69.18 & 0.074 \\ 
WS-3D-Lane  \cite{ai2022ws} & 26.7 & RTX 3080Ti & 77.02 & 84.51 & 70.75 & 0.058 \\ 
Anchor3DLane  \cite{huang2023anchor3dlane} & \textcolor{black}{148} & \textcolor{black}{RTX 3080Ti} & 74.87 & 80.85 & 69.71 & 0.060 \\ 
Efficient Transformer  \cite{chen2023efficient} & 72 & Tesla V100 & \textbf{80.84} & 84.50 & 77.48 & 0.056 \\
GroupLane  \cite{li2023grouplane} & 38.2 & RTX 2080 & 80.73 & 82.56 & \textbf{78.90} & 0.053 \\
LATR  \cite{luo2023latr} & \textcolor{black}{14} & \textcolor{black}{RTX 3080Ti} & 80.59 & \textbf{86.12} & 75.73 & \textbf{0.052} \\
DecoupleLane \cite{han2023decoupling} & - & - & 75.07 & 81.19 & 69.26 & 0.062 \\
CurveFormer++ \cite{bai2024curveformer++} & 11.4 & RTX 3090 & 77.85 & 81.06 & 74.89 & 0.084 \\
\hline
\end{tabular}
\caption{The performance of 3D lane detection on the ONCE-3DLane dataset. CD error represents the Chamfer distance between the predicted lane line and the ground truth.}
\label{table3}
\end{table*}

\section{PERFORMANCE EVALUATION OF 3D LANE DETECTION}
\label{performance_evaluation}
This section will discuss the performance evaluation of monocular 3D lane detection models. Herein, we explain the evaluation metrics, different types of objective functions, analyze the computational complexity, and finally provide quantitative comparisons of various models. The nomenclature of the used notations is given in Table \ref{tab:notations}.

First, we present the visualization results of 3D lane line detection. As some algorithms are not open-sourced, we have only performed visualization tests on the ApolloSim dataset and OpenLane dataset using some open-source algorithms. These algorithms have been trained on the ApolloSim dataset and OpenLane dataset, and the visualization results are shown in Fig. \ref{vis_compare_sim} and Fig. \ref{vis_compare_real}, where the red lines represent the predicted lane lines, and the blue lines represent the ground truth lane lines.
Next, we will introduce the evaluation metrics, the loss function used for training the algorithms, as well as the quantitative test results for 3D lane detection on public datasets, including performance metrics and a comparison of processing delays.

\subsection{Evaluation Metrics for 3D Lane Detection}
Building only predictive \textcolor{black}{monocular} 3D lane detection models is not a wise and trustworthy decision for safe AD unless it is tested on unseen data. Most models evaluate their performance on a disjoint set of the same dataset used for training, i.e., the test data remains new to the training model. 
Deep learning models for monocular 3D lane detection are evaluated using some general metrics to assess the best results based on true values. For monocular 3D lane detection tasks, different types of evaluation metrics are available for these tasks, which we review next as follows:

\subsubsection{Accuracy}
\begin{equation}
Accuracy =   \frac{N_{TP} + N_{TN}}{N_{TP} + N_{FP}  + N_{TN}  + N_{FN}} 
\end{equation}

\subsubsection{Recall}
\begin{equation}
Recall =  \frac{N_{TP}}{N_{TP} + N_{FN}}
\end{equation}

\subsubsection{Precision}
\begin{equation}
Precision =  \frac{N_{TP}}{N_{TP}+N_{FP}}
\end{equation}

\subsubsection{F-Score}
\begin{equation}
F-score = (1+\beta^{2}) \times \frac{Precision * Recall}{\beta^{2} \times Precision + Recall}
\end{equation}

\subsubsection{AP}
\begin{equation}
AP = \sum_{1}^{N}p(k)\Delta r(k)
\end{equation}

\subsubsection{X Error}
\begin{equation}
X_{Error} = \frac{1}{N} \sum_{i=1}^{N} \sqrt{(x_i - \hat{x_i})^2}
\end{equation}

\subsubsection{Z Error}
\begin{equation}
Z_{Error} = \frac{1}{N} \sum_{i=1}^{N} \sqrt{(z_i - \hat{z_i})^2}
\end{equation}


\textcolor{black}{
\subsubsection{CD Error}
\begin{equation}
        CD = \frac{1}{|P|} \sum\limits_{p \in P} \min\limits_{q \in Q } || p - q||^{2} + 
             \frac{1}{|Q|} \sum\limits_{q \in Q} \min\limits_{p \in P } || q - p||^{2}
\end{equation}
}


\subsection{Loss Functions for 3D Lane Detection}
In the task of monocular 3D lane detection, the common basic loss functions include the following:
\subsubsection{MSE loss}
This loss is one of the most commonly used loss functions, which calculates the squared difference between the model's predicted values and the true values, and then takes the average. Its mathematical expression is:

\begin{equation}
  L_{MSE} = \frac{1}{N} \sum_{i=1}^{N} (y_i - \hat{y}_i)^2 
\end{equation}

\subsubsection{MAE Loss}
This loss is another commonly used loss function, which calculates the absolute difference between the model's predicted values and the true values, and then takes the average. Its mathematical expression is:
\begin{equation}
    L_{MAE} = \frac{1}{N} \sum_{i=1}^{N} |y_i - \hat{y}_i|
\end{equation}

\subsubsection{Huber Loss}
Huber Loss combines the advantages of MSE and MAE, making it more robust to outliers. Its mathematical expression is:
\begin{equation}
L_{Huber} = \begin{cases} 
\frac{1}{2}(y - \hat{y})^2, & \text{if } |y - \hat{y}| \leq \delta \\ 
\delta (|y - \hat{y}| - \frac{1}{2}\delta), & \text{otherwise} 
\end{cases}
\end{equation}

\subsubsection{Cross-Entropy Loss}
Cross-Entropy is commonly used for classification tasks but can also be applied to regression tasks. In lane detection, the problem can be transformed into a classification task by determining whether a pixel belongs to a lane based on its position. Its mathematical expression is:
\begin{equation}
L_{CE} = -\frac{1}{N} \sum_{i=1}^{N} \sum_{c=1}^{C} y_{ic} \log(p_{ic})
\end{equation}


\subsubsection{Binary Cross-Entropy Loss}
Binary Cross-Entropy loss is commonly used for training binary classification tasks, aiming to minimize the loss function to improve the accuracy of the model's predictions on binary classification samples. It is widely applied in deep learning for tasks such as image classification, text classification, and segmentation. Its mathematical expression is:
\begin{equation}
L_{BCE} = - \frac{1}{N} \sum_{i=1}^{N} \left[ y_i \log(p_i) + (1 - y_i) \log(1 - p_i) \right]
\end{equation}


\subsubsection{Focal Loss}
Focal Loss is a loss function designed to address the class imbalance problem commonly encountered in tasks such as object detection or semantic segmentation, where the number of examples from one class greatly outnumbers the number of examples from another. This class imbalance can lead to models being biased towards the majority class, resulting in poor performance, especially on the minority class.
\begin{equation}
L_{Focal} = - (1 - p_t)^\gamma \log(p_t)
\end{equation}

\subsubsection{IoU Loss}
IoU Loss is based on Intersection over Union (IoU) and is used in object detection and segmentation tasks to measure the overlap between the predicted regions by the model and the ground truth regions.
\begin{equation}
\textcolor{black}{L_{IoU} = 1 - \frac{|S \cap \hat{S}|}{|S \cup \hat{S}|} }
\end{equation}

Different methods utilize specific loss functions differently, but essentially, most are variations or combinations of the basic loss functions mentioned above. Additionally, the Hungarian algorithm \cite{enwiki:1212340485} is often used to match predicted lanes with ground truth lanes.



\subsection{Quantitative Analysis of Monocular 3D Lane Detection Models}
\begin{figure*}[h]
    \setlength{\abovecaptionskip}{0pt}
    \setlength{\belowcaptionskip}{0pt}
    \centering
    \includegraphics[width=0.78\linewidth]{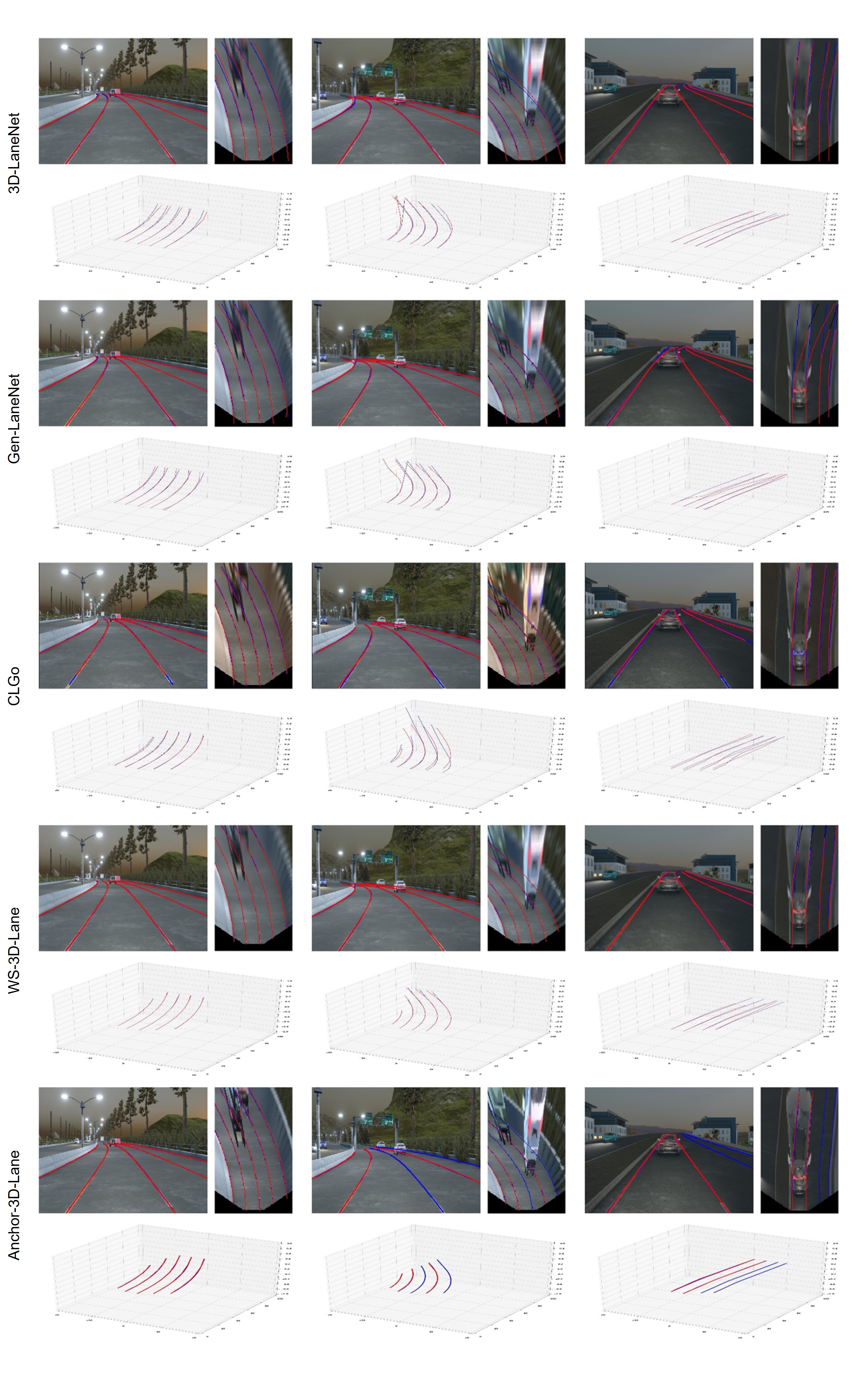}
    \captionsetup{font={small}}
    \caption{
    Visualization results of several open-source algorithms on the ApolloSim dataset \cite{guo2020gen}, with 3D-LaneNet \cite{efrat20203d}, Gen-LaneNet \cite{guo2020gen}, CLGo \cite{liu2022learning}, WS-3D-Lane \cite{ai2022ws}, and Anchor3DLane \cite{huang2023anchor3dlane} algorithms displayed from top to bottom. In the images, the blue lines represent the ground truth of the lane lines, while the red lines represent the lane lines predicted by the algorithms.}
    \label{vis_compare_sim}
\end{figure*}

\begin{figure*}[h]
    \setlength{\abovecaptionskip}{0pt}
    \setlength{\belowcaptionskip}{0pt}
    \centering
    \includegraphics[width=0.90\linewidth]{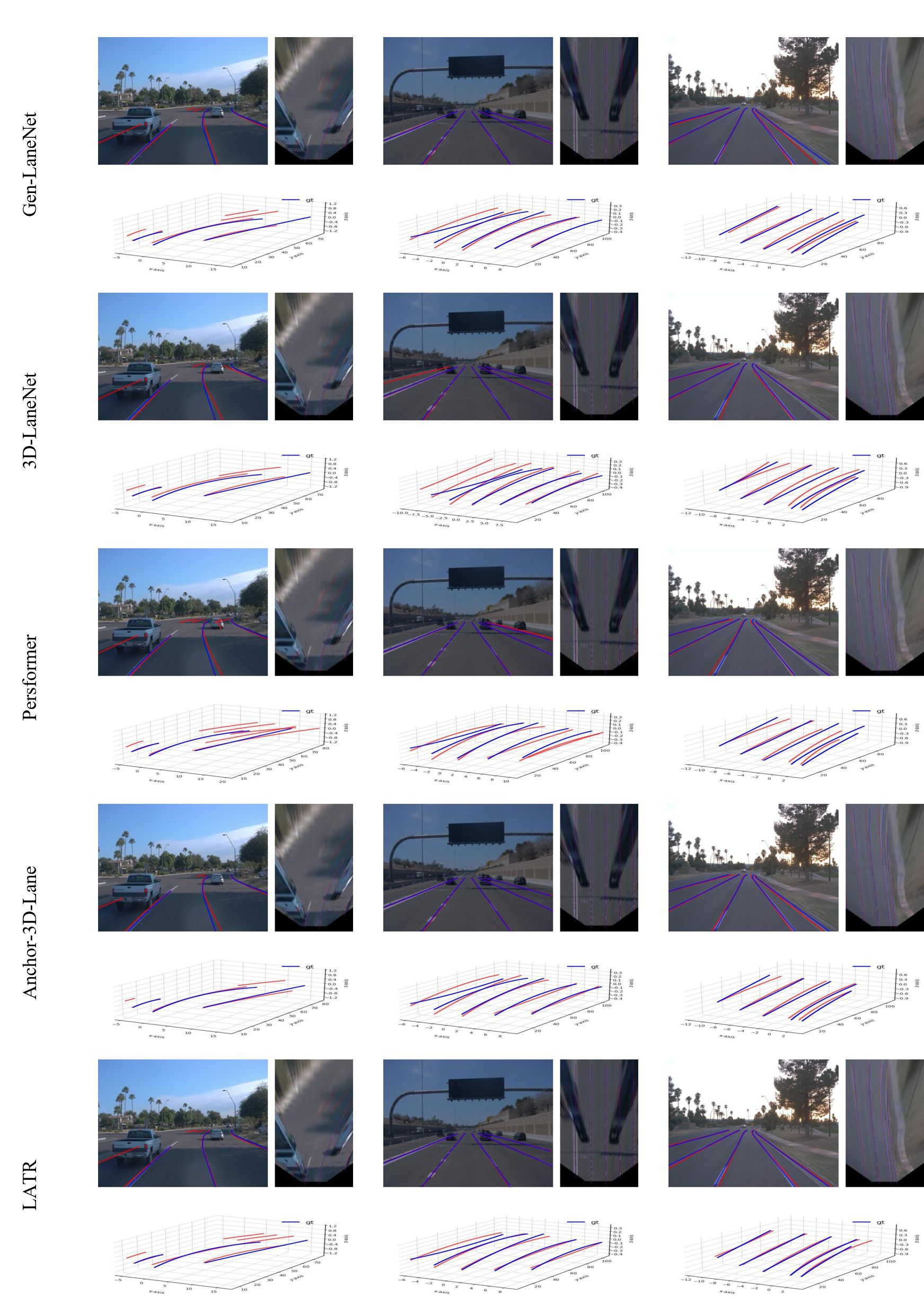}
    \captionsetup{font={small}}
    \caption{
    \textcolor{black}{Visualization results of several open-source algorithms on the OpenLane dataset \cite{chen2022persformer}, with Gen-LaneNet \cite{guo2020gen}, 3D-LaneNet \cite{efrat20203d}, Persformer \cite{chen2022persformer}, Anchor3DLane \cite{huang2023anchor3dlane} and LATR \cite{luo2023latr} algorithms displayed from top to bottom. In the images, the blue lines represent the ground truth of the lane lines, while the red lines represent the lane lines predicted by the algorithms.}}
    \label{vis_compare_real}
\end{figure*}

This section elaborates on the quantitative empirical analysis of monocular 3D lane detection methods surveyed in this paper. 
For quantitative assessment, we utilize four evaluation metrics to examine the performance of each monocular 3D lane detection method on the ApolloSim dataset: \textit{AP}, \textit{F-Score}, \textit{x error} and \textit{z error}, and we report the results in Table \ref{performance_comparison}. 
On the OpenLane dataset, we evaluated the \textit{F-Score} of each model, as shown in Table \ref{table2}. 
On the ONCE-3DLane dataset, we evaluate four metrics, namely: \textit{F-Score}, \textit{Precision}, \textit{Recall}, and \textit{CD error}, the results are reported om Table \ref{table3}. The qualitative visualization results on the ApolloSim dataset and OpenLane dataset are shown in Fig. \ref{vis_compare_sim} and Fig. \ref{vis_compare_real}.

Additionally, computational efficiency is taken into account by reporting the Frames Per Second (FPS) achievable by each method during the inference process.
The overall runtime of these models is reported in Table \ref{performance_comparison}, \ref{table2}, and \ref{table3}. 
\textcolor{black}{
When evaluating computational efficiency, we use our unified computing platform to measure latency for open-source algorithms. For methods that are not open-source, we directly utilize the reported latency and corresponding computing platform details provided in the original papers.
Our experimental platform's CPU configuration includes an Intel(R) Core i9-12900K CPU processor running on the Ubuntu 20.04 operating system, while the GPU used in the experiments is an NVIDIA GeForce RTX 3080Ti GPU with 12 GB of graphics memory.
}
\textcolor{black}{From the table above, it can be observed that all methods have the potential to operate in real-time within AD systems, thus affirmatively addressing the question we raised in Section \ref{questions}. Among these methods, BEV-LaneDet \cite{wang2023bev}, Anchor3DLane \cite{huang2023anchor3dlane}, and CLGo \cite{liu2022learning} demonstrate very fast processing speeds, which is advantageous for practical deployment. Methods like LATR \cite{luo2023latr}, D-3DLD \cite{kim2023d}, and WS-3D-Lane \cite{ai2022ws}, while meeting real-time requirements, exhibit higher processing latencies and would benefit from acceleration techniques, such as utilizing TensorRT for optimized deployment.}

\section{DATASETS}
\label{datasetes}
In deep learning-based vision tasks, an equally important component is the dataset. In this section, we introduce the datasets used for monocular 3D lane detection tasks. Some of these datasets are open-sourced and popular, frequently used by the community, while others are only described in papers and are not open-sourced. Whether open-source or proprietary, we have compiled a detailed table to provide a more intuitive overview of these datasets, as presented in Table \ref{tab:dataset}.

\subsection{Apollo 3D Lane Synthetic Dataset}

The Apollo 3D Lane Synthetic Dataset \cite{guo2020gen} is a robust synthetic dataset comprising 10,500 frames of high-resolution 1080 × 1920 monocular RGB images, built using the Unity 3D engine. Each frame is accompanied by 
corresponding 3D lane labels and camera pitch data. 
It based on Silicon Valley in the United States, spans a wide array of environments, including highways, urban areas, residential neighborhoods, and downtown settings.

The dataset's images encapsulate a broad range of daytime and weather conditions, various traffic/obstacle scenarios, and different road surface qualities, thereby providing the dataset with a high degree of diversity and realism.
The dataset is segmented into three distinct scene categories: Balanced Scenes, Rarely Observed Scenes, and Scenes with Visual Variations. Balanced Scenes serve as a comprehensive and unbiased dataset for benchmarking standard driving scenarios. Rarely Observed Scenes are used for testing an algorithm's adaptability to rarely encountered situations involving complex urban maps with drastic elevation changes and sharp turns. Scenes with Visual Variations are designed to assess how algorithms perform under different illumination conditions by excluding specific daytime periods during training and focusing on them during testing.
The camera’s intrinsic parameters in the dataset are fixed, with the camera heights ranging between 1.4 to 1.8 meters and pitch angles spanning from 0 to 10 degrees.

\subsection{OpenLane}

OpenLane \cite{chen2022persformer} stands as the first large-scale, real-world 3D lane detection dataset, boasting over 200,000 frames and 880,000 meticulously annotated lanes. Built on the foundation of the influential Waymo Open dataset, OpenLane utilizes the same data format, evaluation pipeline, and a 10Hz sampling rate, courtesy of 64-beam LiDARs, over a 20-second span.
This dataset provides exhaustive details for each frame, inclusive of camera intrinsics and extrinsics, along with lane categories, which encompass 14 distinct types such as white dashed lines and road curbsides. Almost 90\% of the total lanes are constituted by double yellow solid lanes, and single white solid and dashed lanes. The OpenLane dataset epitomizes a real-world scenario, duly highlighting the long-tail distribution problem.
OpenLane includes all lanes in a frame, even accommodating those in the opposite direction, provided there is no dividing curbside. 
Due to complex lane topologies, such as intersections and roundabouts, a single frame can accommodate up to 24 lanes.
Approximately 25\% of the frames harbor more than six lanes, surpassing the maximum in most current lane datasets. In addition to this, the dataset provides annotations for scene tags such as weather and locations, along with the closest-in-path object (CIPO) - defined as the most pertinent target with respect to the ego vehicle. This ancillary data proves invaluable for subsequent modules in planning and control, and not merely perception.
The 3D ground truth of OpenLane is synthesized using LiDAR, thus ensuring high precision and accuracy. The dataset is split between a training set containing 157,000 images and a validation set comprising 39,000 images.

\begin{table*}[h]
\centering
\caption{A brief introduction to the datasets in monocular 3D lane line detection, including their source papers, parent dataset, data type, resolution, data size, the year in which they were created, availability, and the link to their open-source repositories.}
\begin{tabular}{c c c c c c c c}
\hline
\textbf{Dataset} & \textbf{Parent Dataset} & \textbf{Data Type} & \textbf{Resolution} & \textbf{Total Images} & \textbf{Year} & \textbf{Availability} & \textbf{Source} \\  \hline
ApolloSim \cite{guo2020gen}       & -                       & Synthetic (Unity 3D) & 1920 × 1080       & 10.5K              & 2020 & \ding{52} & \scriptsize \textbf{\href{https://github.com/yuliangguo/3D_Lane_Synthetic_Dataset}{link}} \\ 
OpenLane  \cite{chen2022persformer}       & Waymo  \cite{sun2020scalability}   & Real-world            & 1920 × 1280       & 200K               & 2022 & \ding{52} & \scriptsize \textbf{\href{https://github.com/OpenDriveLab/OpenLane}{link}}\\ 
ONCE-3DLanes  \cite{yan2022once}   & ONCE    \cite{mao2021one}                & Real-world            & 1920 × 1020       & 211K               & 2022 & \ding{52} & \scriptsize \textbf{\href{https://github.com/once-3dlanes/once_3dlanes_benchmark}{link}}\\ 
Synthetic-3D-Lanes \cite{garnett20193d} & -                      & Synthetic (Blender)  & 360 × 480         & 306K               & 2019 & \ding{56} & - \\ 
Real-world 3D-Lanes \cite{garnett20193d} & -                & Real-world            &  /  & 85K                & 2019 & \ding{56} & -\\ 
LLAMAS-3DLD   \cite{kim2023d}   & LLAMAS    \cite{behrendt2019unsupervised}             & Real-world            & 1276 × 717        & 6.6K               & 2023 & \ding{56} & -\\ 
Pandaset-3DLD   \cite{kim2023d} & Pandaset    \cite{xiao2021pandaset}            & Real-world            & 1920 × 1080       & 6K                 & 2023 & \ding{56} & -\\ \hline
\end{tabular}
\label{tab:dataset}
\end{table*}

\subsection{ONCE-3DLanes}

The ONCE-3DLanes dataset \cite{yan2022once} manifests as another practical 3D lane detection dataset, meticulously derived from the ONCE autonomous driving repository. This collection encompasses 211,000 visuals captured by a forward-facing camera, complemented by their matching LiDAR point cloud data.
Showcasing a wide spectrum of scenes across varied times of day and weather settings, such as sunlit, overcast, and rainy conditions, the dataset casts light on multiple terrains like city centers, residential areas, highways, bridges, and tunnels. This diversity fortifies the dataset as a crucial resource for developing and validating resilient 3D lane detection models under an assortment of real-world scenarios.
It is organized into three segments: 3,000 scenes for validation, 8,000 scenes for testing, and the residual 5,000 scenes reserved for training. The training component is supplemented with an extra 200,000 unlabeled scenarios to thoroughly harness the raw data. While the dataset supplies intrinsic camera parameters, it omits extrinsic camera parameters.

\subsection{Other Datasets}
In \cite{garnett20193d}, it presents two distinct datasets: the Synthetic-3D-Lanes dataset and the 3D-Lanes dataset. The Synthetic-3D-Lanes dataset, created via the open-source graphics engine, Blender, consists of 300K training and 5K testing examples, each featuring a 360×480 pixel image alongside associated ground truth parameters such as 3D lanes, camera height, and pitch. Exhibiting substantial diversity in lane topography, object placement, and scene rendering, this dataset provides a valuable resource for method development and ablation studies.
Complementarily, the 3D-Lanes dataset is a real-world ground truth labeled data collection, compiled through the utilization of a multi-sensor setup - a forward-looking camera, a Velodine HDL32 lidar scanner, and a high-precision IMU. 
This dataset consists of six independent driving records, each recorded on a different road segment, totaling nearly two hours of driving time.
With the aid of Lidar and IMU data, aggregated lidar top view images were generated and used alongside a semi-manual annotation tool to establish the ground truth. In total, 85,000 images were annotated, of which 1,000, derived from a separate drive, were designated as the test set, with the remainder serving as the training set. The 3D-Lanes dataset proves instrumental in validating the transferability of the proposed approach to real-world data and enabling qualitative analysis.
While the Synthetic-3D-Lanes dataset is made available to the research community, the real-world 3D-Lanes dataset remains proprietary and is not publicly accessible. It's worth noting that despite its availability, the Synthetic-3D-Lanes dataset has not seen wide adoption for benchmarking in subsequent studies in the field.

\textcolor{black}{
As shown in Table \ref{tab:dataset}, although there are a number of datasets currently available in the field of 3D lane detection, only a few are open-source. Regarding the question raised in section \ref{questions}, we believe that while these datasets do include some complex scenarios, they remain insufficient. For example, there is still a lack of datasets under nighttime and extreme weather conditions. To promote further development in this field, more challenging data should be added in the future. In addition to collecting real-world data, using the latest text-to-image and text-to-video generation technologies also offers promising possibilities for gathering data on challenging scenarios.
}


\section{3D LANE DETECTION IN AD: CHALLENGES AND DIRECTIONS}
\label{challenges_and_directions}

The datasets introduced above encompass a variety of publicly available road scenes. Current mainstream research primarily focuses on favorable daytime scenes that are suitable for and conducive to 3D lane detection, with sufficient lighting and favorable weather conditions. Many car companies and original equipment manufacturers in the industry possess a large volume of data, however, they are reluctant to share it publicly due to concerns related to intellectual property, industrial competition, and General Data Protection Regulations (GDPR). As a result, the lack of sufficient annotated data for accurately understanding dynamic weather conditions, such as nighttime, smoggy weather, and edge cases, remains a challenging task in AD research.

This research area is among the challenges that the community has not yet adequately addressed. In this section, we present critical viewpoints on the current state of 3D lane detection in AD, summarize a series of challenges, and provide research direction recommendations to help the community make further progress and effectively overcome these challenges.

\subsection{Open Challenges}
Although significant research has been conducted by researchers in the field of AD, and the AD industry is flourishing, there are still open challenges that require researchers' attention to achieve fully intelligent AD. These challenges have been individually discussed with support from relevant literature:
\textcolor{black}{\subsubsection{Coarse-Structured Information}}
Most of the datasets introduced in the literature for 3D lane detection in AD are recorded in normal and well-structured infrastructures of advanced cities. The currently developed deep learning models may achieve the best results over structured datasets, but they generalize poorly in many unstructured environments. This issue of AD demands further attention in terms of data collection, as well as the inclusion of new and effective representation mechanisms in deep learning models.
\subsubsection{Uncertainty-Aware Decisions}
A largely overlooked aspect in lane detection and AD decision-making is the confidence with which models make predictions on input data. However, the confidence of model outputs plays a crucial role in ensuring the safety of AD. The inherently uncertain nature of the vehicular surroundings seems to have not convinced the community to delve into this matter, as the current methodological trends focus solely on predictive scores. Fortunately, confidence estimation has recently gained attention in the community \cite{arnez2020comparison, michelmore2018evaluating}. Nevertheless, elements from evidential deep learning \cite{sensoy2018evidential}, Bayesian formulations of deep neural networks \cite{kendall2017uncertainties}, simpler mechanisms to approximate the output confidence of neural networks (e.g., Monte Carlo dropout \cite{gal2016dropout} or ensembles \cite{lakshminarayanan2017simple}), and other assorted methods for uncertainty quantification \cite{abdar2021review} should be progressively incorporated as an additional yet crucial criterion for decision making. When dealing with complex environments, where there is a lack of data that fully represents all possible scenes, the model outputs a significant amount of epistemic uncertainty. Without considering confidence as an additional factor for AD or with current studies focused solely on predictive and/or computational efficiency aspects, there is no guarantee that the emerging 3D lane detection models in the scientific community will be practically useful and transferable to the industry.
\subsubsection{Weakly Supervised Learning Strategies}
In current deep learning-based models, the majority rely on fully supervised learning strategies, which have high demands for labeled data. This is particularly challenging in the field of 3D lane detection, as general visual sensor data lacks depth information. It is difficult to simply assign 3D information to lanes based on images alone, and alternative sensors such as LiDAR are required to obtain the 3D lane information. This leads to the costly and labor-intensive task of annotating 3D lane data. Fortunately, both the academic and industrial communities have recognized this issue, and weakly supervised learning strategies have received extensive research and attention in the field of deep learning. However, in the specific branch of 3D lane detection, there is currently limited research focusing on weakly supervised learning strategies. 
If we can effectively leverage self-supervised/weakly supervised learning strategies, it would significantly reduce data collection costs and allow for more training data to enhance the performance of 3D lane detection algorithms, thereby further advancing the AD industry.

\subsection{Future Directions}

\subsubsection{\textbf{Video-Based 3D Lane Detection for Autonomous Driving}}

Drawing from the advancements in video-based object detection \cite{wu2019sequence}, semantic segmentation \cite{wang2021swiftnet,maninis2018video,hu2018videomatch}, and 2D lane detection \cite{Jin2023RecursiveVL,zhang2021vil,zhang2018end}, it is evident that incorporating video-based techniques significantly improves the precision and dependability of 3D lane detection systems. The core advantage of video-based methods is their capacity to harness temporal data, providing a dynamic view that static images lack. This dynamic perspective is particularly vital in comprehending and predicting complex driving situations in 3D spaces, where the intricacies of lane placement and vehicular interactions intensify. Methods like Recursive Video Lane Detection (RVLD) \cite{Jin2023RecursiveVL} demonstrate the power of video in capturing ongoing lane shifts and variations over time, a feature immensely beneficial for 3D modeling accuracy. 
Additionally, incorporating video data into these systems enhances our understanding of spatial dynamics in driving environments, which is crucial for 3D lane detection.
By incorporating sophisticated deep learning techniques used in video-based object detection and semantic segmentation, future iterations of 3D lane detection systems could attain advanced spatial awareness, significantly enhancing the navigation capabilities and safety of autonomous vehicles.

\subsubsection{\textbf{Hybrid Approaches and Multimodality}}

The progression of multi-modal 3D lane detection technologies for autonomous vehicles is greatly accelerated by the incorporation of various sensory inputs, such as cameras, LiDAR, and radar. This integration marks a promising avenue for overcoming the challenges faced by existing camera-reliant systems. This approach, underscored by the successes in multi-modal 3D object detection and semantic segmentation \cite{feng2020deep}, leverages the complementary strengths of each sensor type to enhance detection accuracy and reliability, particularly in challenging environmental and complex driving scenarios \cite{chen2019progressive, liang2019multi}. Reflecting on pioneering models such as \cite{bai2018deep} and \cite{luo2022m}, which have effectively utilized multi-sensor inputs to refine lane boundary estimations and demonstrated robust performance against occlusions and variable lighting conditions, the potential for significant advancements in lane detection technology is evident. The future trajectory in this field should emphasize the exploration of advanced data fusion methods, meticulous sensor calibration, and synchronization techniques, alongside the exploitation of emerging technologies like edge computing for real-time multi-modal data processing.

\subsubsection{\textbf{Active Learning and Incremental Learning}}
Active learning \cite{ren2021survey} in machine learning allows a model to adapt and learn over time. This learning process takes place when the model encounters new data during testing and after deployment, making it especially valuable in real-world environments.
In real-world environments, vehicles may encounter randomly occurring unfamiliar scenes and lane topology, which may require artificial intelligence models to make decisions for further actions such as braking or accelerating to achieve sensible driving operations. Therefore, lane detection techniques should allow for an interactive approach to handle various types of scenes and lane topologies, involving human annotators to label unlabeled data instances and human involvement in the training process \cite{khan2022pmal}. There are different types of active learning techniques, such as membership query synthesis \cite{kelner2020learning}, where synthetic data is generated and the parameters of the synthetic data can be adjusted based on the structure of the data \cite{ahmed2022aaqal}, which originates from the underlying species of the dataset. On the other hand, the ability of 3D lane detection models to incrementally update their captured knowledge with new data is crucial for their sustainability and continuous improvement. We anticipate that both these capabilities of 3D lane detection models in road understanding will become increasingly important in future research.

\subsubsection{\textbf{Adverse Weather Conditions}}

The progress in developing camera-based 3D lane detection systems for AD is notably hindered by adverse weather conditions, which significantly impair visibility. Events like heavy rainfall, fog, snow, and dust storms can severely impact the functionality of these systems \cite{zhang2023perception}. The core issue stems from the compromised quality of visual data necessary for the precise detection and segmentation of lane markings, resulting in a decline in reliability and an increase in the likelihood of false negatives or positives. This diminishment in the efficacy of such systems not only elevates safety hazards but also constrains the operational scope of autonomous vehicles. However, recent breakthroughs in object detection and segmentation, as illustrated in research works such as \cite{sakaridis2021acdc} and \cite{9184996}, reveal avenues for enhancing 3D lane detection amidst challenging weather conditions. These studies put forth the advantages of employing deep learning algorithms trained on datasets that include a wide array of adverse weather instances, showcasing the significance of effective data augmentation, domain adaptation tailored to specific conditions, and the employment of semantic segmentation techniques. Employing these approaches, the capabilities of camera-based detection systems can be substantially improved to interpret lane markings accurately and ensure safe navigation even under poor visibility, laying down an optimistic pathway for ongoing research and development in the field of AD technology.

\begin{figure*}[t]
    \setlength{\abovecaptionskip}{0pt}
    \setlength{\belowcaptionskip}{0pt}
    \centering
    \includegraphics[width=1.0\linewidth]{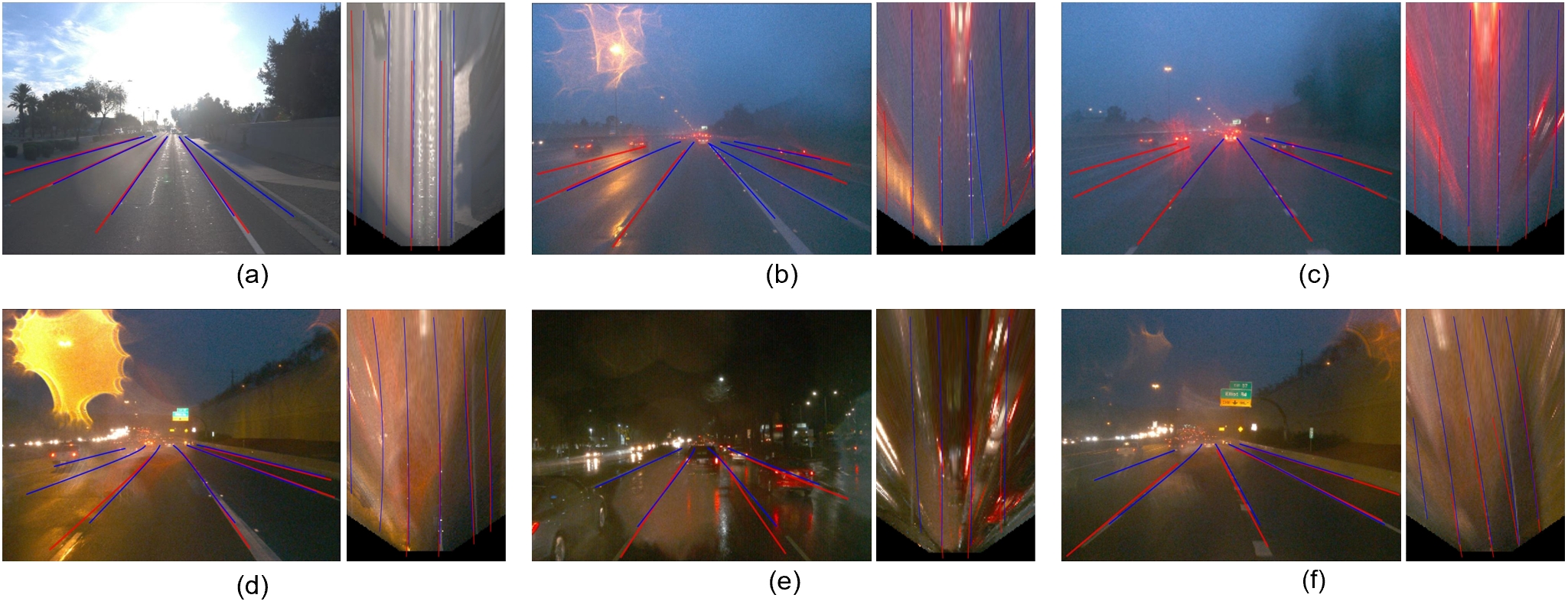}
    \captionsetup{font={small}}
    \caption{
    The test results of the current open-source state-of-the-art LATR \cite{luo2023latr} algorithm under extreme weather conditions or photography scenarios are depicted, with red lines indicating the algorithm's predicted lane lines and blue lines representing the ground truth lane lines. (a) and (d) show scenes with strong exposure during the day and at night, respectively; figures (b) and (e) depict rainy scenes during the day and night; and figures (c) and (f) show foggy scenes during the day and night. From these test scenarios under harsh conditions, it is evident that the current algorithm performs poorly in detecting lane lines, indicating that there is substantial room for improvement. Additionally, there is a lack of datasets for training and testing under such severe conditions.}
    \label{compare_bad_condition}
\end{figure*}

\subsubsection{\textbf{3D Lane Detection with Large Language Model}}

The advent of (LLMs), such as ChatGPT \cite{wu2023brief}, have transformed the field of artificial general intelligence (AGI), demonstrating their remarkable zero-shot abilities in tackling a variety of natural language processing (NLP) tasks using customized user prompts or language instructions.
Computer vision encompasses a set of unique challenges and concepts that stand apart from those found in NLP. Vision foundation models typically adhere to a pre-training and subsequent fine-tuning process \cite{wang2023internimage, chen2022vision, su2023towards, wang2022image, fang2023eva, tao2023siamese}, which, while effective, entails significant additional costs for adaptation to a range of downstream applications. Techniques such as multi-task unification \cite{radford2018improving, wang2022ofa, alayrac2022flamingo, wang2022git} aim to endow systems with a broad set of capabilities, yet they often fail to transcend the constraints of predetermined tasks, leaving a noticeable capability deficit in open-ended tasks when juxtaposed with LLMs. The advent of visual prompt tuning \cite{jia2022visual, yao2021cpt, zang2022unified, wang2023images} provides a novel method for delineating specific vision tasks, like object detection, instance segmentation, and pose estimation, through the use of visual masking. Nonetheless, there is currently no work that combines LLMs with 3D lane line detection. As LLMs become more widespread and their capabilities continue to improve, research into LLM-based lane line detection presents an interesting and promising avenue for future exploration.

\subsubsection{\textbf{Towards More Accurate and Efficient 3D Lane Detection Methods for Autonomous Driving}}
\label{qqqqq}
The qualitative performance of current 3D lane detection techniques is shown in Table \ref{performance_comparison}. We can observe that only a few methods manage to strike a balance between model accuracy and inference latency. The experimental results of these methods indicate the need for further improvements to alleviate computational burdens while maintaining their unparalleled performance. 
Additionally, we select some challenging data from the 3D lane detection dataset and test the performance of the 3D lane line detection algorithm on these challenging data samples. 
However, the performance of the algorithm is not satisfactory, as depicted in Fig. \ref{compare_bad_condition}. 
\textcolor{black}{This also answers the question we raised in section \ref{questions}.}
Improving the algorithm's detection performance in extreme weather conditions is also crucial. 
Moreover, the reported time complexities in Table \ref{performance_comparison},\ref{table2} and \ref{table3} suggest that some methods can achieve real-time execution when deployed on GPU devices. Nevertheless, considering the constrained computational resources in AD systems, the focus of 3D lane detection methods should also shift towards computational complexity.

\subsubsection{\textbf{3D Lane Detection based on Event Cameras}}
RGB cameras are limited by their imaging principles, resulting in poor image quality in high-speed or low-light scenes.
Fortunately, event cameras can overcome this limitation. 
Event cameras are visual sensors with high temporal resolution, high dynamic range, low latency, and low energy consumption \cite{enwiki:1202151981}.
Unlike traditional cameras that capture images based on the intensity and color of light, event cameras capture images based on changes in light intensity \cite{gallego2020event,gao2022vector}. Therefore, event cameras can capture images in low-light scenes as long as there are changes in light intensity. 
Currently, there is limited research on 3D lane detection based on event cameras. We believe that there is significant and extensive research potential in the field of 3D lane detection using event cameras, including the development of datasets specifically for 3D lane detection with event cameras and the design of algorithms tailored for 3D lane detection using event cameras alone or in combination with RGB cameras.

\subsubsection{\textbf{Uncertainty-Aware 3D Lane Detection}}
Over the past few years, DNNs have garnered significant success in numerous computer vision assignments, solidifying their status as indispensable instruments for proficient automated perception. Despite consistently delivering exceptional results across different benchmarks and tasks, 
There are still several significant hurdles that need to be overcome before widespread implementation. Among the most common and well-known critiques of DNNs is their susceptibility to unreliable performance when confronted with shifts in data distribution levels, highlighting the urgent need to rectify this limitation \cite{franchi2022muad}.
Currently, the majority of deep learning models provide deterministic outputs, giving a single result. However, in real-world driving scenarios, it is desirable for models to provide uncertainty estimates for their predictions \cite{mcallister2017concrete,kendall2017uncertainties,ma2023self}. 
Downstream decision-making modules can then use this uncertainty information to make more reasonable and safe driving commands. For example, in the context of 3D lane detection, if the model outputs a lane position with high uncertainty, we should approach the model's detection result with skepticism, and a conservative driving style should be adopted. Conversely, if the model's output has low uncertainty, we can have confidence in the algorithm's prediction and make more confident driving decisions.

\section{CONCLUSION}
\label{Conclusion}
Vision sensors are essential in autonomous vehicles, as they significantly influence the decision-making process. As one of the fastest-evolving areas in recent years, computer vision techniques analyze data from these sensors to extract crucial information, including traffic light detection, traffic sign recognition, drivable area delineation, and 3D obstacle perception. With advancements in sensor technology, algorithmic sophistication, and computational power, leveraging vision sensor data for autonomous vehicle perception has garnered growing attention. Monocular-based 3D lane detection, for instance, uses a single camera image to identify lane line positions within the 3D environment, integrating valuable depth information. Depth knowledge of lane lines is essential for making safe, comfortable decisions and planning maneuvers. Although 3D lane information can also be derived from other sensors, such as LiDAR, vision sensors remain invaluable in AD for their cost-effectiveness and rich, structured color information.

Monocular image-based 3D lane detection has been extensively studied in AD, however, comprehensive analysis and synthesis of this research are scarce in the current literature. In this survey, we systematically review existing lane detection methods, introduce available 3D lane detection datasets, and discuss the performance of these methods across public datasets. Additionally, we examine the challenges and limitations that current 3D lane detection approaches encounter. Our findings highlight that monocular image-based 3D lane detection remains an evolving field with notable limitations, which we discuss in detail along with corresponding recommendations and outlooks. We cover foundational work in deep learning models, outline their structural hierarchy for 3D lane detection tasks, and address the specific challenges encountered by each model category. Furthermore, we analyze performance evaluation strategies, loss functions, and widely used datasets in AD relevant to 3D lane detection models. Finally, we summarize open challenges and potential future research directions, providing reference to key baseline studies from recent literature to foster further progress in this area.

Finally, it is clear that experts in the intelligent transportation systems field are continuously advancing strategies for 3D lane detection to fully harness data from vision sensors. Current research predominantly focuses on enhancing model accuracy via neural networks or by exploring innovative network architectures. However, addressing additional challenges remains essential to achieve reliable, trustworthy, and safe AD. Specifically, in 3D lane detection, these challenges demand the development of more robust models capable of anticipating lane occlusions, processing coarse-structured information, and issuing risk alerts. 
Moreover, existing 3D lane detection models largely depend on supervised learning, which necessitates high-quality labeled data—a labor-intensive and time-consuming requirement. Thus, exploring alternative methodologies such as self-supervised or weakly supervised learning presents a valuable and promising path for further development in this area. Effectively leveraging these opportunities could significantly propel research in intelligent transportation systems, advancing 3D lane detection to new heights. This progress would enable more effective deployment of autonomous vehicles in real-world settings, supporting safer, more reliable, and comfortable travel and logistics solutions.

\renewcommand*{\bibfont}{\footnotesize}
\printbibliography



\end{document}